%% file: main.tex
\documentclass[sigconf,9pt]{acmart}

\AtBeginDocument{%
  }

\acmYear{2023}\copyrightyear{2023}
\setcopyright{acmlicensed}
\acmConference{Arxiv}
\acmBooktitle{Preprint}

\renewcommand\footnotetextcopyrightpermission[1]{}
\usepackage{enumitem}
\usepackage{booktabs,siunitx}
\usepackage{caption}
\usepackage{algorithm}
\usepackage{ulem}
\usepackage[noend]{algpseudocode}
\usepackage{makecell}

\usepackage{multicol}
\usepackage{multirow}
\usepackage{subfigure}
\usepackage{bbding}

\begin{document}


\title{DC-CCL: Device-Cloud Collaborative Controlled Learning for Large Vision Models}


\author{Yucheng Ding}
\email{yc.ding@sjtu.edu.cn}
\affiliation{%
  \institution{Shanghai Jiao Tong University}
  \country{China}
}

\author{Chaoyue Niu}
\email{rvince@sjtu.edu.cn}
\affiliation{%
  \institution{Shanghai Jiao Tong University}
  \country{China}
}

\author{Fan Wu}
\email{wu-fan@sjtu.edu.cn}
\affiliation{%
  \institution{Shanghai Jiao Tong University}
  \country{China}
}

\author{Shaojie Tang}
\email{shaojie.tang@utdallas.edu}
\affiliation{%
  \institution{University of Texas at Dallas}
  \country{USA}
}

\author{Chengfei Lyu}
\email{chengfei.lcf@alibaba-inc.com}
\affiliation{%
  \institution{Alibaba Group}
  \country{China}
}

\author{Guihai Chen}
\email{gchen@cs.sjtu.edu.cn}
\affiliation{%
  \institution{Shanghai Jiao Tong University}
  \country{China}
}

\begin{abstract}
Many large vision models have been deployed on the cloud for real-time services. Meanwhile, fresh samples are continuously generated on the served mobile device. How to leverage the device-side samples to improve the cloud-side large model becomes a practical requirement, but falls into the dilemma of no raw sample up-link and no large model down-link. Specifically, the user may opt out of sharing raw samples with the cloud due to the concern of privacy or communication overhead, while the size of some large vision models far exceeds the mobile device's runtime capacity. In this work, we propose a device-cloud collaborative controlled learning framework, called DC-CCL, enabling a cloud-side large vision model that cannot be directly deployed on the mobile device to still benefit from the device-side local samples. In particular, DC-CCL vertically splits the base model into two submodels, one large submodel for learning from the cloud-side samples and the other small submodel for learning from the device-side samples and performing device-cloud knowledge fusion. Nevertheless, on-device training of the small submodel requires the output of the cloud-side large submodel to compute the desired gradients. DC-CCL thus introduces a light-weight model to mimic the large cloud-side submodel with knowledge distillation, which can be offloaded to the mobile device to control its small submodel's optimization direction. Given the decoupling nature of two submodels in collaborative learning, DC-CCL also allows the cloud to take a pre-trained model and the mobile device to take another model with a different backbone architecture. We extensively evaluate DC-CCL over 5 public datasets and 6 common models, demonstrating its effectiveness and efficiency in approaching the performance of ideally leveraging the large model, as well as its remarkable advantage over the baseline of exploiting only the cloud-side samples or adopting only a device-affordable small model.

\end{abstract}
\keywords{device-cloud collaborative learning, mobile computer vision applications, large vision models, vertical model decoupling}

\maketitle

\input{01Introduction}
\input{02Background}

\input{03-1Principle}
\input{03-2Framework}
\input{04Evaluation}
\input{05RelatedWork}

\input{06Conclusion}

\input{07Appendix}

\clearpage
\bibliographystyle{ACM-Reference-Format}
\bibliography{dc-ccl-ref}

\end{document}

%% file: 01Introduction.tex
\section{Introduction}

Nowadays, more and more deep vision models are deployed on the cloud server to provide mobile APP users with diverse intelligent services, such as image recognition (e.g., in Google Lens and Google Photos), which categorizes the photos taken by mobile device users in daily life; livestreaming highlight recognition (e.g., in TikTok, Taobao Live, and Weixin Channels Live), which identifies the key time points of a streamer in introducing attractive information; and video analytics (e.g., in YouTube, YouKu, and Kuaishou), which adds tags, extracts topics, and identifies key events from the published videos for viewers. In the meanwhile, during the usage of these mobile APPs, fresh samples with user feedback are continuously generated on each served mobile device.

To learn from the device-side samples and exploit the natural advantages of mobile devices being close to users and data sources, some work explored the new paradigm of device-cloud collaborative learning \cite{proc:osdi22:walle}. One line of work \cite{yao_kdd21,yan_kdd22} relied on the raw sample up-link from the mobile device to the cloud, which is feasible only when data privacy and communication overhead do not pose major concerns. Another line of work resorted to model down-link. The celebrated framework is cross-device federated learning (FL) \cite{fed}, which requires mobile devices to download the full affordable model and train it from scratch, while the cloud periodically aggregates the model updates. Other work on efficient on-device learning \cite{quan,prune,mobilenet,tinytl}
considered the case that the original model breaks the resource limit of the mobile device, enabled model offloading by reducing model size and optimizing model structure, and further allowed cloud-to-device knowledge transfer and device-side adaptation. 

Different from existing work, we consider how the cloud-side model benefits from the fresh device-side samples (i.e., the need to transfer the device-side knowledge back to the cloud and fuse it with the cloud-side knowledge) without raw sample up-link and without large model down-link in the computer vision (CV) scenarios. 

On the one hand, the local vision samples of each mobile device, which contain sensitive information, cannot be uploaded to the cloud to protect user privacy. In addition, compared with each user's daily typed words in natural language processing (NLP) scenarios and daily behavior data in recommender systems, which are in the size of KB, the size of 100 images with $224\times224$ pixels is roughly 14MB, and the size of a 5 minutes-long, 1080p, and 8Mbps video is roughly 300MB. Therefore, it is also communication efficient for each mobile device to avoid uploading the vision samples. 

\begin{figure}[!t]
\centering
\includegraphics[width=\columnwidth]{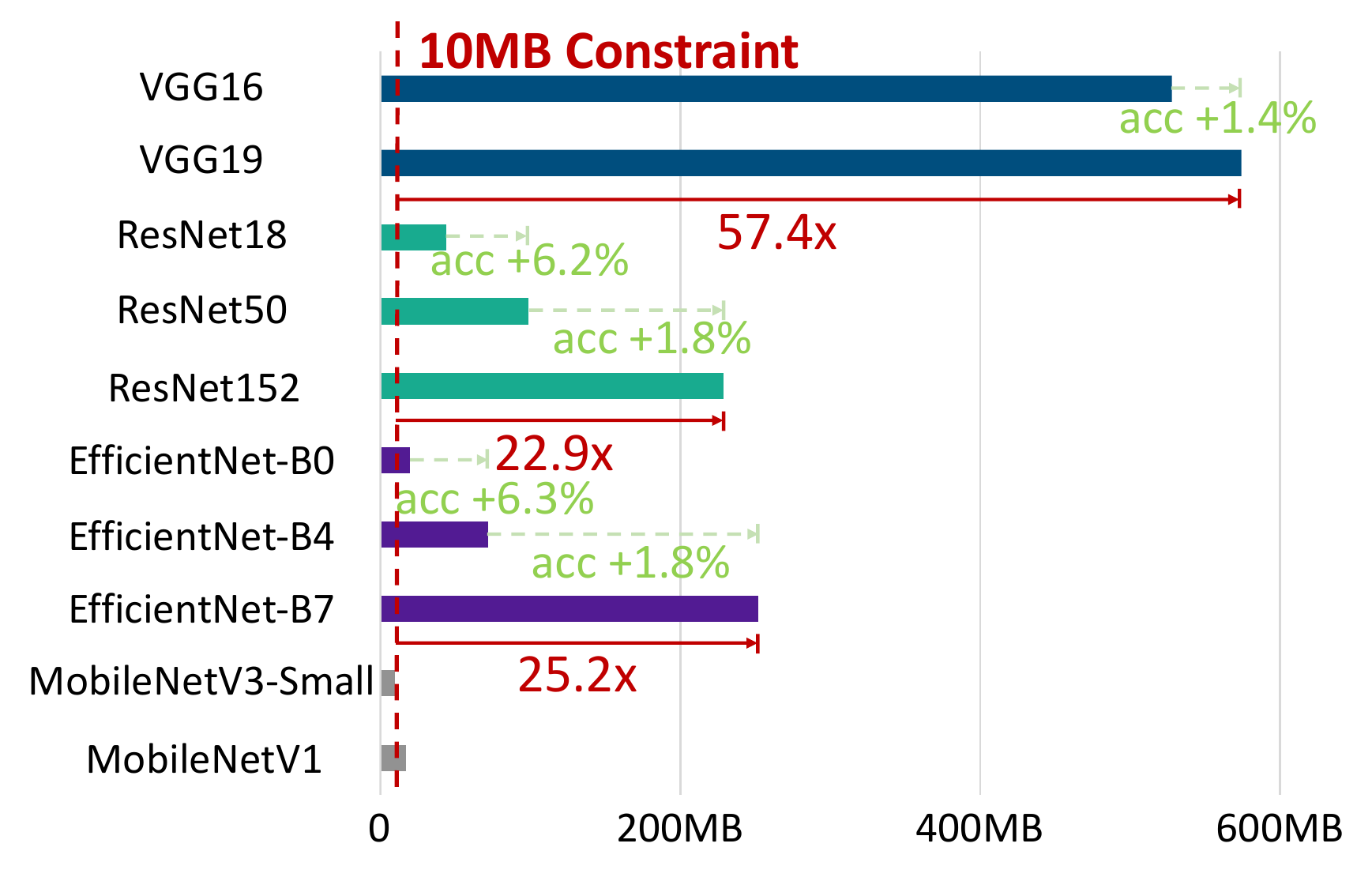}
\caption{Sizes of some classical CNNs on ImageNet-1K.}
\label{cnn_sizes}
\end{figure}

On the other hand, the cloud-side vision model normally has a large size of parameters for a strong recognition ability, which, however, may exceed the runtime capacity of mobile APPs. According to the deployment practice in industry, the size of the model to be trained in a mobile APP should be strictly no larger than 10MB, preventing APP crash and poor user experience. In contrast, as depicted in Figure \ref{cnn_sizes}, the sizes of several mainstream convolutional neural networks (CNNs) designed for the cloud-based image classification task with high accuracy (e.g., VGG19, ResNet152, and EfficientNet-B7) are larger than 200MB. Furthermore, different from NLP and recommendation models with a dominant sparse embedding layer \cite{niu_mobicom20}, the vision models follow dense connection patterns and do not have much structural redundancy. As a result, some cloud-based large vision models cannot be structurally compressed with a high compression ratio and a low accuracy loss such that they can be offloaded to the mobile device for local training and further facilitate transferring the device-side knowledge back to the cloud for fusion.  

Given the practical requirement and challenges above, we propose a new Device-Cloud Collaborative Controlled Learning framework, denoted as DC-CCL for short. The key of DC-CCL is to decouple the on-device learning process from the dependence on the full model. The motivating intuition is that training over the mobile device's few fresh samples needs to update only a small part of the full model, not requiring most of the model parameters for learning over the cloud-side, large-scale samples. Therefore, DC-CCL vertically splits the base model with an existing backbone network architecture into: (1) a large cloud-side submodel for learning the cloud-side knowledge as regular; and (2) a small device-cloud co-submodel for learning device-side knowledge and fusing it with the cloud-side knowledge. Due to the decoupling feature, two submodels can take different backbone architectures. In addition, the optimization process is correspondingly divided into two decoupled phases: (1) training the cloud-side submodel purely on the cloud from scratch or directly taking a pre-trained model; and (2) training the device-cloud co-submodel with the collaboration of the cloud and the mobile device in a data parallelism way. However, on-device training of the co-submodel still requires the outputs of the cloud-side submodel, which is unaffordable by the mobile device. To deal with this problem, DC-CCL introduces a light-weight model to mimic the cloud-side submodel with knowledge distillation and offloads it to the mobile device, helping compute the desired gradient of the co-submodel and controlling its optimization direction. 

We summarize the key contributions as follows: 
\begin{itemize}
    \item DC-CCL is the first device-cloud collaborative learning framework that enables cloud-side large vision models to learn from device-side samples without the raw sample up-link and the large model down-link.
    \item DC-CCL decouples on-device learning from the dependence on the cloud-side full model, by updating only a small submodel vertically split out and further introducing a light-weight model, which can approximate the remaining large submodel's outputs, to control local optimization.
    \item Evaluation results over several public datasets and different models reveal that (1) the device-side model size is 4.32\% -- 27.86\% of the cloud-side model size and can scale down or up to balance efficiency and accuracy in DC-CCL; and (2) DC-CCL improves the accuracy by 3.52\% -- 41.32\%, compared with the baselines of training only with the cloud-side samples or adopting a small device-affordable model.
\end{itemize}
 

%% file: 02Background.tex
\section{Problem Formulation}

\begin{figure}[!t]
\centering
\includegraphics[width=0.95\columnwidth]{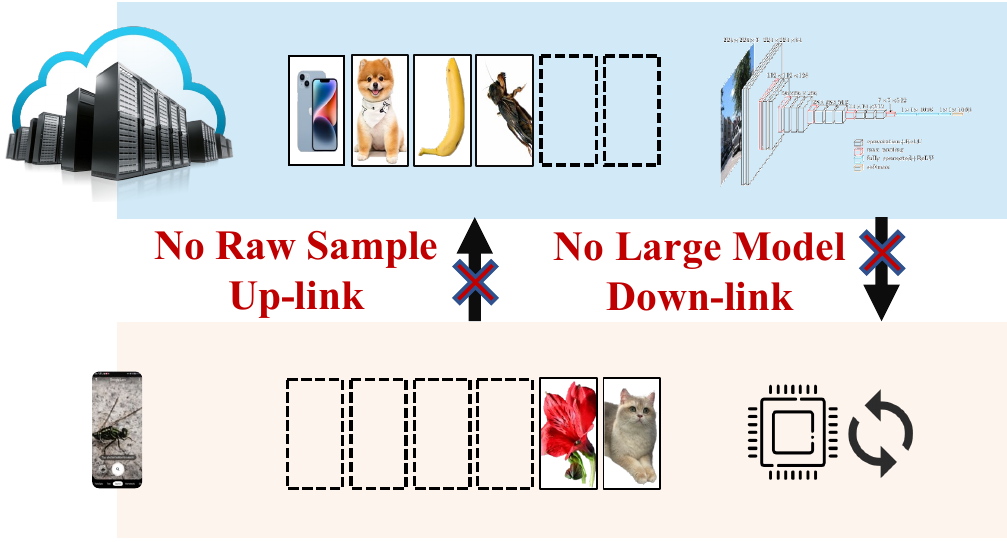}
\caption{An illustration of the considered device-cloud collaborative learning scenario.}
\label{problem_setting}
\end{figure}

\begin{figure*}[!t]
\centering
\subfigure[Base Model]{
\includegraphics[width=0.29\linewidth]{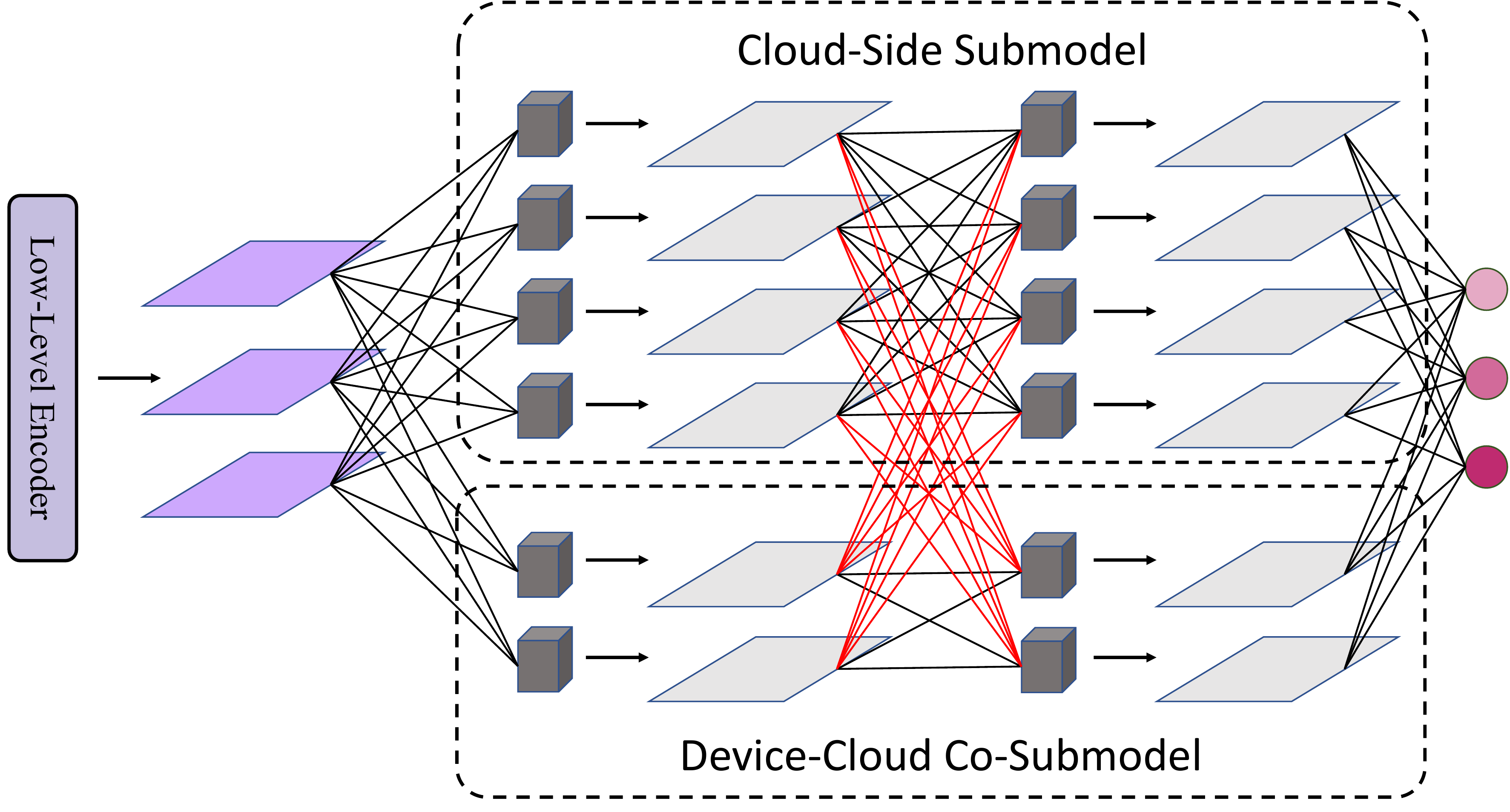}
\label{dense_model_arc}
}
\subfigure[Decoupled Model]{
\includegraphics[width=0.33\linewidth]{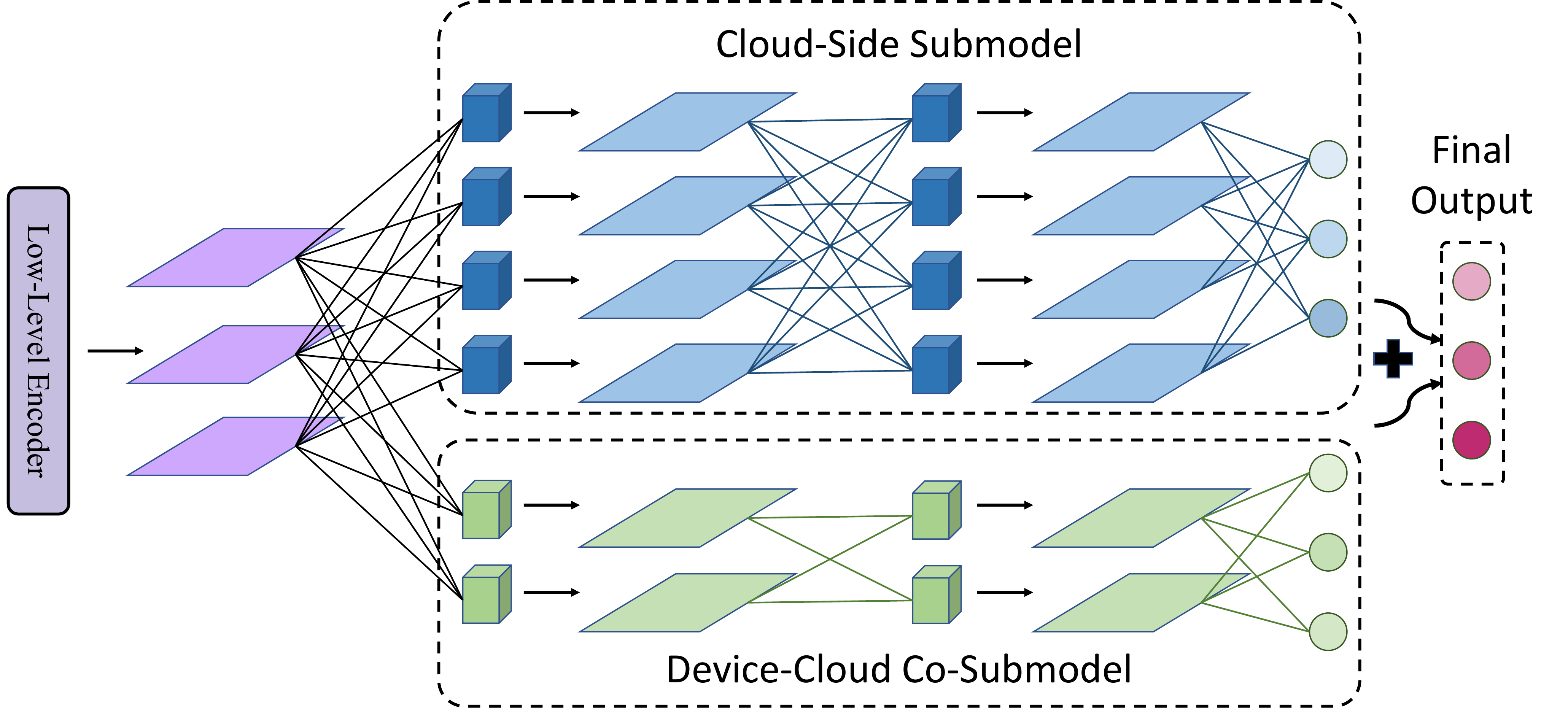}
\label{iso_model_arc}
}
\subfigure[Device-Side Model]{
\includegraphics[width=0.33\linewidth]{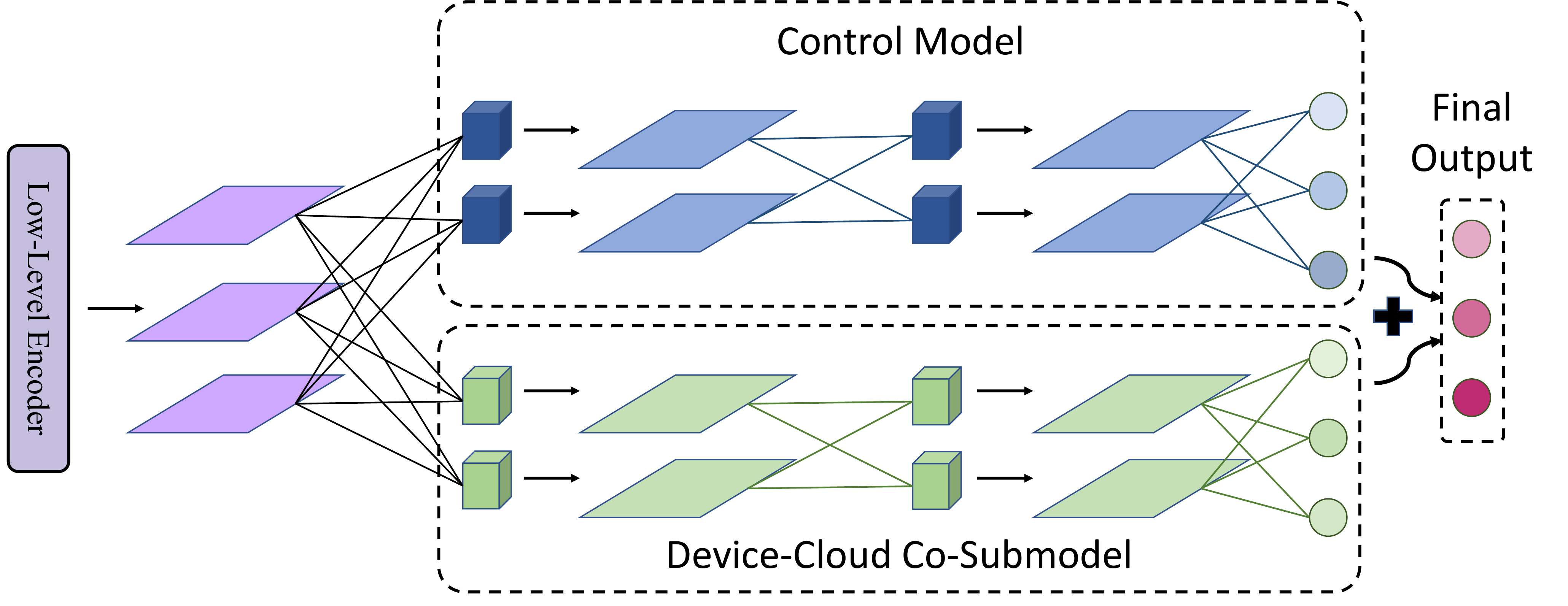}
\label{device_model_arc}
}
\caption{Illustrations of a base model with an existing backbone network architecture, the corresponding decoupled model on the cloud, and the device-side model. The forward/backward processes of the cloud-side submodel and the device-cloud co-submodel are mutually dependent in the base model, but are isolated in the decoupled model. The device-side model contains the co-submodel and a light-weight control model that mimics the cloud-side submodel. The parallelograms denote feature maps, and the cubes denote filters.}
\end{figure*}

In this section, we describe the considered scenario, the problem, and the motivations in detail. 

As shown in Figure \ref{problem_setting}, we focus on CV tasks with two practical considerations on the sample up-link from a mobile device to the cloud and the model down-link from the cloud to the mobile device. In particular, (1) users can generate or collect some fresh vision samples and store them locally on their mobile devices. In addition, a user may be reluctant to upload (part of) his/her samples to the cloud in practice. The concerns mainly come from privacy and security (e.g., sensitive personal photos) as well as communication overhead (e.g., because the size of vision data is normally larger than the size of data in other modals, and the up-link of a mobile device tends to pose the bottleneck); and (2) the cloud maintains a large model with a strong recognition ability. Nevertheless, the model size far exceeds the runtime capacity of a mobile device (i.e., 10MB for mobile APPs in industry). and cannot be offloaded for on-device learning; otherwise, the limits of memory and CPU will be broken, leading to service crash and degrading user experience. 

Given the disconnections of the sample up-link and the large model down-link between the cloud and the mobile device, we consider how to enable the large model on the cloud to benefit from the local samples on the mobile device, without uploading the local samples and without offloading the large model. Such a problem is well motivated, because the mainstream cloud-based learning framework optimizes the large model only over the cloud-side samples, but fails to leverage device-side samples, while more data can effectively improve the model's generalization ability. Meanwhile, this problem is also atypical in the isolation setting of the large model and the samples, whereas existing learning frameworks require the model and the samples to be put together (e.g., on the cloud under the cloud-based learning framework or on each mobile device under the FL framework).

%% file: 03-1Principle.tex
\section{Design Rationale}\label{ADR}

In this section, we demystify the design rationale and the key insights of DC-CCL. 

For a large base model using an existing backbone network architecture, we view its optimization process over the samples on the cloud and on the mobile device as a full learning task. The key is to free on-device learning from the dependence on the full base model. Intuitively, the size of device-side samples is small, and there is no need to use a large model. Therefore, as shown in Figure~\ref{dense_model_arc}, a natural idea is to split out most of the base model, called cloud-side submodel, be responsible for the large-scale samples on the cloud, and let the rest small part, called device-cloud co-submodel, learn from device-side samples and further fuse device-side new knowledge with cloud-side knowledge. Based on this intuition, we propose to divide the full learning task of the base model into two subtasks: one subtask for first training the cloud-side submodel over the samples on the cloud; and the other subtask for the mobile device and the cloud to collaboratively train the co-submodel using their respective samples in a data parallelism way. 

However, the subtask (i.e., the forward pass and the backward propagation) of device-cloud co-submodel depends on the cloud-side submodel, as depicted by the red connections in Figure~\ref{dense_model_arc}. This implies that the mobile device still needs to use the unaffordable cloud-side submodel. To decouple two submodels, we propose to cut off the connections between their neurons, as shown in Figure \ref{iso_model_arc}. From the direction of network splitting (i.e., being perpendicular to the input layer), we essentially design a vertical splitting method for the base model. The final model output now takes the sum of the submodels' outputs. This indicates that when the mobile device optimizes the co-submodel over its local samples, the training loss and the gradient should be derived based on the sum of the cloud-side submodel output and the co-submodel output. Nevertheless, the cloud-side submodel cannot be offloaded to the mobile device, and its output is absent. To deal with this problem, we introduce a light-weight control model to mimic the cloud-side submodel output with knowledge distillation. As shown in Figure \ref{device_model_arc}, the model deployed on the mobile device includes the device-cloud co-submodel and the control model, which jointly can approximate the desired gradient for local optimization.         

By overviewing the design flow above, DC-CCL starts from the base model with a certain backbone network architecture, vertically splits it into two submodels with the same backbone, and further decouples the learning processes of two submodels by cutting off their neuron connections and introducing a light-weight control model to mimic the large submodel for on-device deployment. Given the isolation feature of training two submodels, DC-CCL also supports that the cloud-side submodel and the device-cloud co-submodel take different backbone network architectures.  

\section{Feasibility Study}\label{ftim}

\begin{figure}[!t] 
\centering 
\begin{minipage}{0.99\columnwidth} 
\centering 
\includegraphics[width=0.85\linewidth]{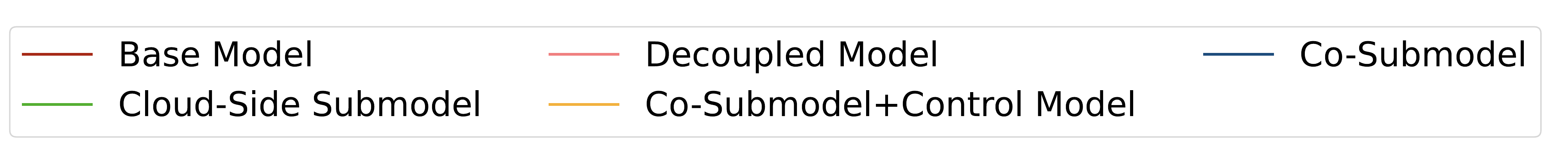}
\end{minipage}
\begin{minipage}{0.99\columnwidth} 
\vspace{-0.1cm}
\centering
\subfigure[Whole Training Process]{
\includegraphics[width=0.45\columnwidth]{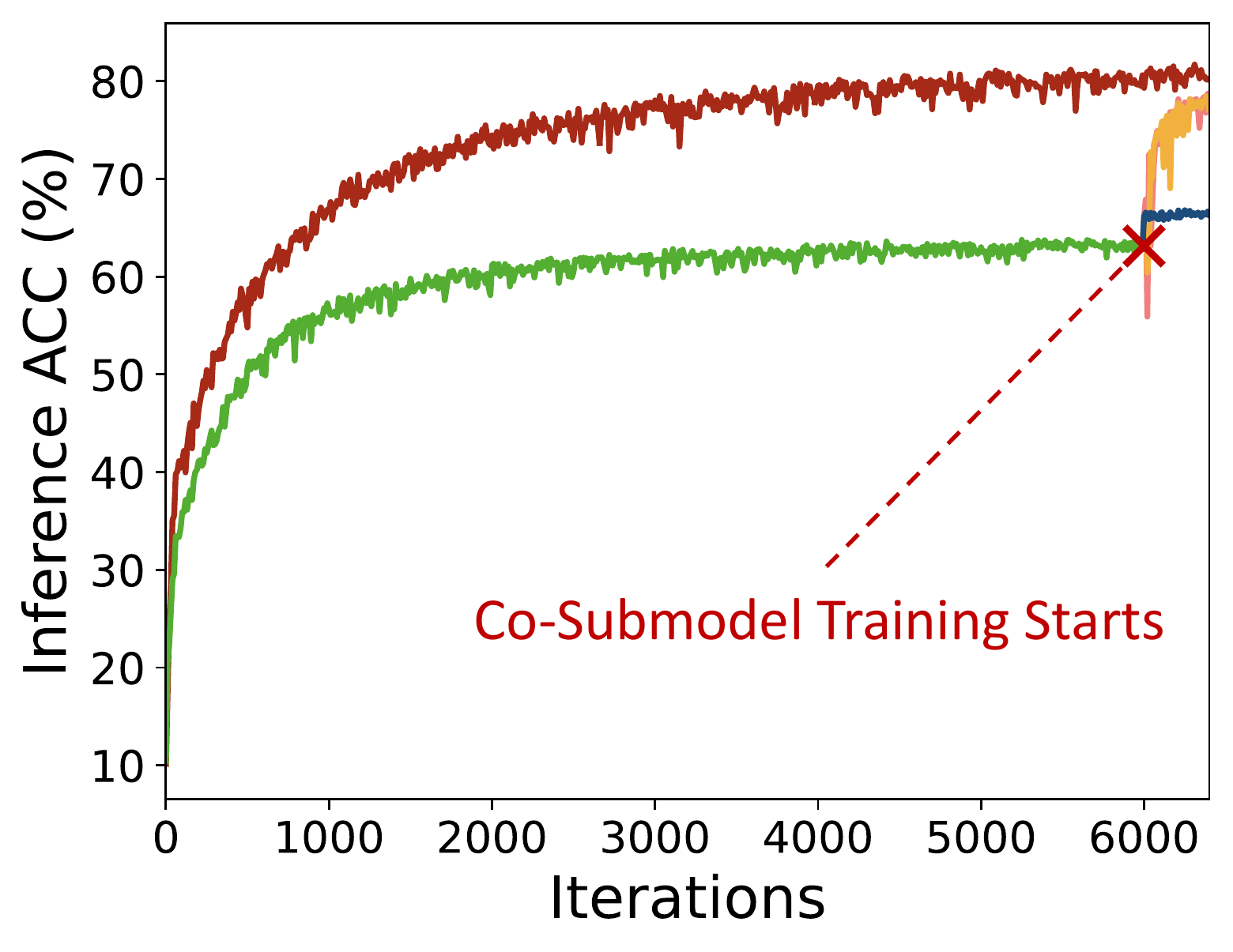}
}
\subfigure[Training Co-Submodel]{
\includegraphics[width=0.47\columnwidth]{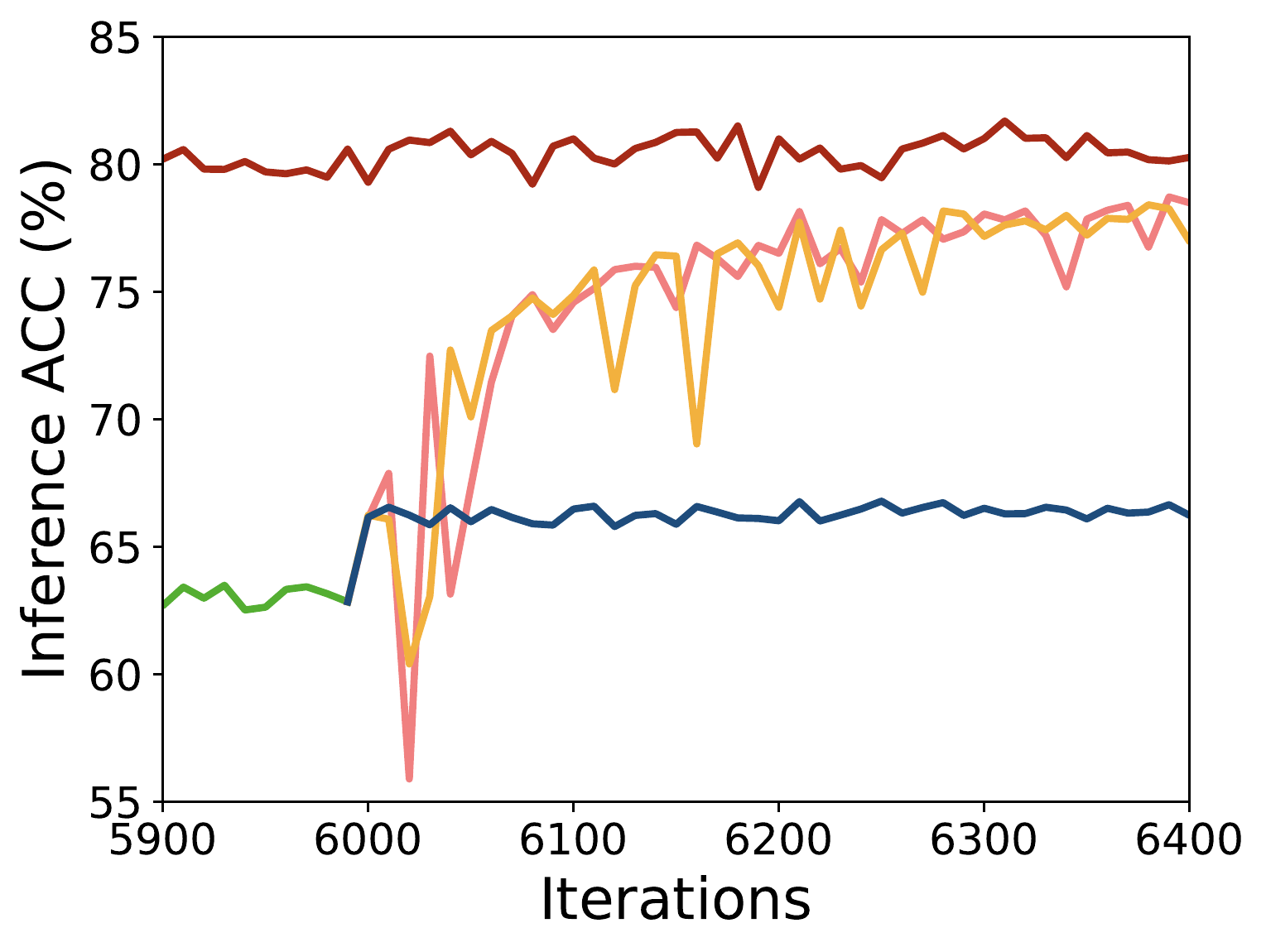}
}
\end{minipage} 
\vspace{-5pt}
\caption{Feasibility study results of model decoupling. Training the device-cloud co-submodel under the correction of the light-weight control model achieves similar accuracy to ideally training the base model.}
\vspace{-10pt}
\label{submodel-acc-val}
\end{figure}

In this section, we study the feasibility of the design rationale. In particular, we validate that through vertical model splitting and submodel learning process decoupling, (1) the device-cloud co-submodel can learn device-side new knowledge and fuse it with cloud-side knowledge; and (2) the device-cloud co-submodel can be trained well with the light-weight control model, not depending on the cloud-side submodel and the full base model any more, but cannot be trained well without the control model.  

We take CIFAR10 dataset and divide it into two subsets for the cloud and the mobile device, covering 8 classes and 2 classes of images, respectively. We use a 5-layer CNN as the base model for image classification, comprised of 4 convolution layers and 1 fully connected layer. We reserve the first convolution layer as a shared low-level encoder and vertically split the other layers into a cloud-side submodel and a device-cloud co-submodel. We let the control model take the same structure as the device-cloud co-submodel. The size of the device-side model, consisting of the low-level encoder, the device-cloud co-submodel, and the control model, is only 5.1\% of the base model. More details about the experimental setups, including model/submodel structures and sizes, are deferred to Table \ref{toy_model} in Appendix \ref{exp_detail}.

\begin{figure}[!t]
    \centering
    \subfigure[Cloud-Side Submodel]{
    \includegraphics[width=0.45\columnwidth]{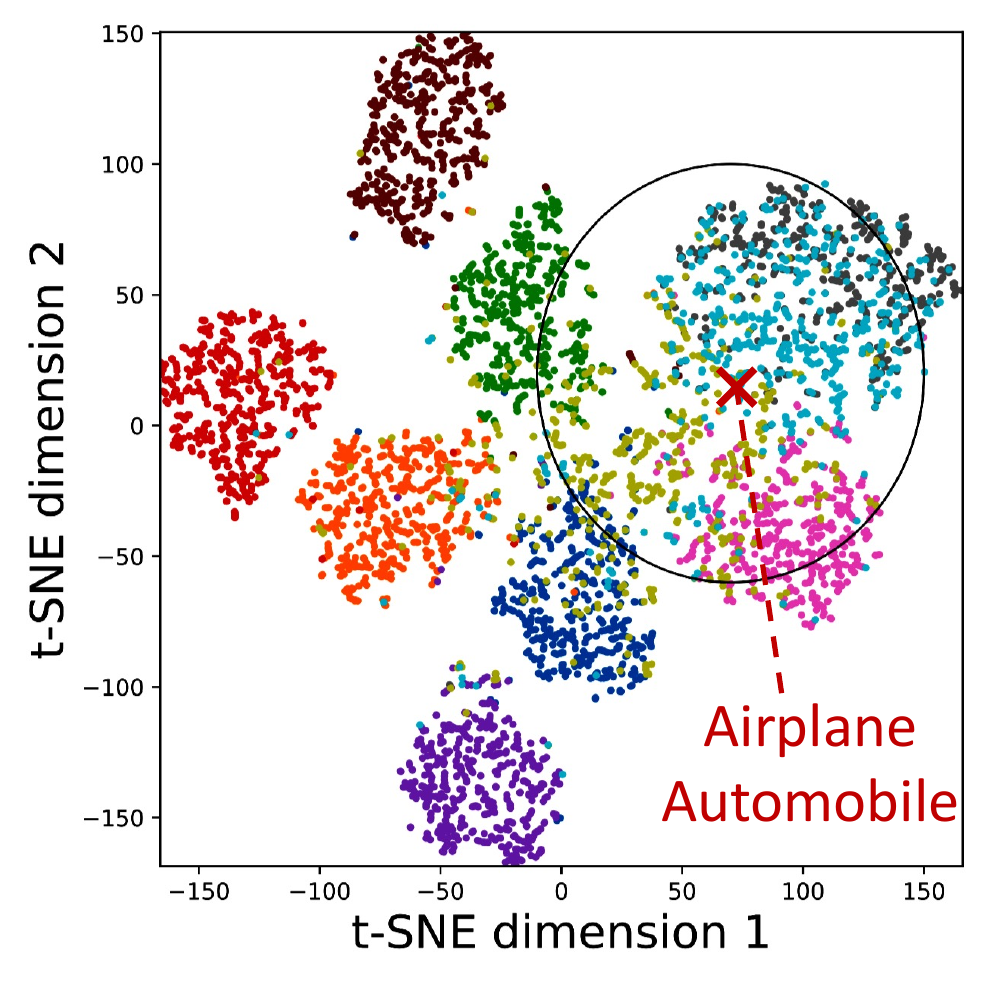}
    }
    \subfigure[Decoupled Model]{
    \includegraphics[width=0.45\columnwidth]{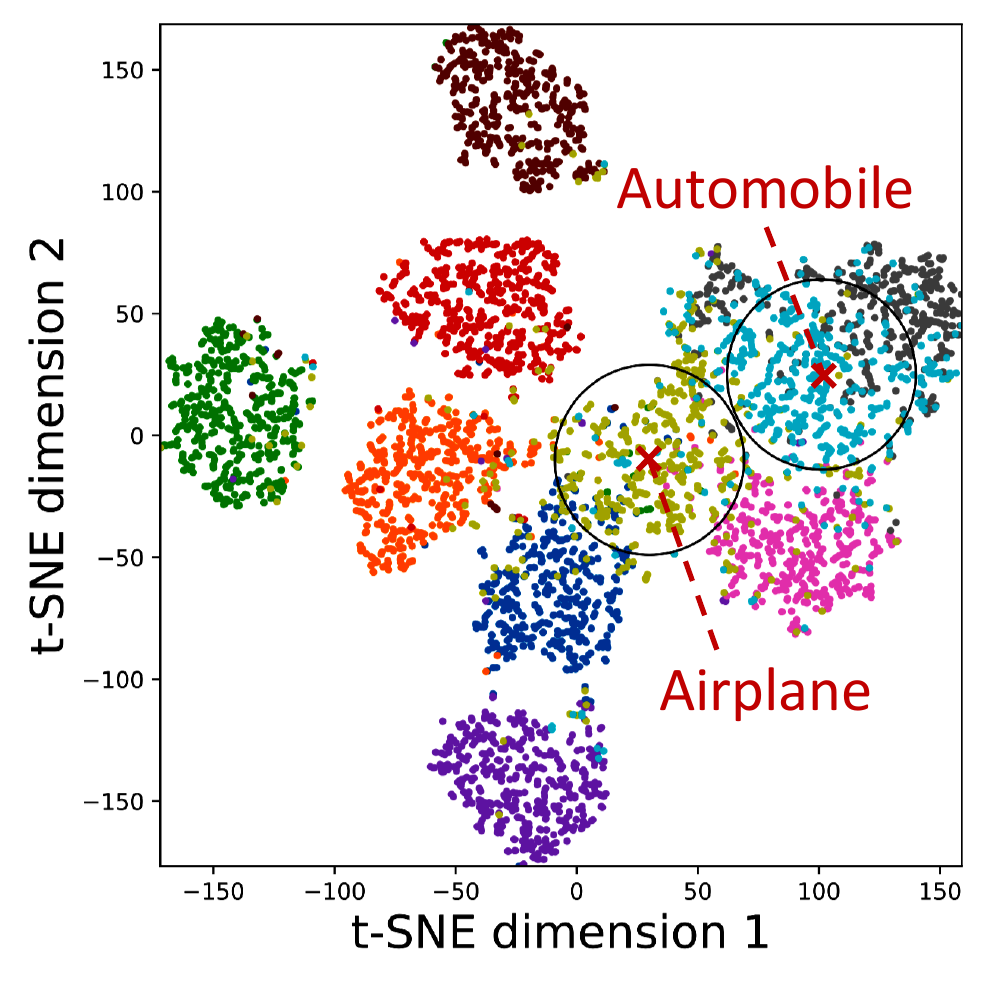}
    }
    \vspace{-0.5em}
    \caption{Difference in recognition ability without and with device-cloud co-submodel.}\label{submodel-acc-val-visual}
\end{figure}

To verify the feasibility of model decoupling, we train the decoupled model in two stages. In the first stage, the shared encoder and the cloud-side submodel are trained over the cloud-side samples of 8 classes and then frozen (i.e., will not be updated in the backward propagation) in the second stage, where only the device-cloud co-submodel is trained over the full dataset. We plot the classification accuracy of the decoupled model in Figure \ref{submodel-acc-val}. We observe that compared with the ideal baseline of training the full base model over the full dataset, the decoupled model can achieve similar accuracy. Intuitively, as depicted in Figure \ref{submodel-acc-val-visual}, the cloud-side submodel learns how to classify the samples on the cloud, while the device-cloud co-submodel is responsible not only for classifying device-side samples but also for distinguishing the device-side samples from the cloud-side ones.

We further verify the feasibility and the necessity of the control model in training the device-cloud co-submodel, when the cloud-side submodel is absent. The key difference from the above experiment lies in the second stage: we first train the light-weight control model to mimic the cloud-side submodel with knowledge distillation over the cloud-side samples; and then replace the cloud-side submodel with the control model. We also remove the control model for ablation study. As shown in Figure \ref{submodel-acc-val}, with the control model, the accuracy can approach the ideal baseline; but without it, the accuracy is roughly 15\% lower than the ideal baseline.

%% file: 03-2Framework.tex
\section{Design Details}

\begin{figure*}[!t]
\centering
\subfigure[Cloud-Based Training and Distillation]{
\includegraphics[width=0.28\linewidth]{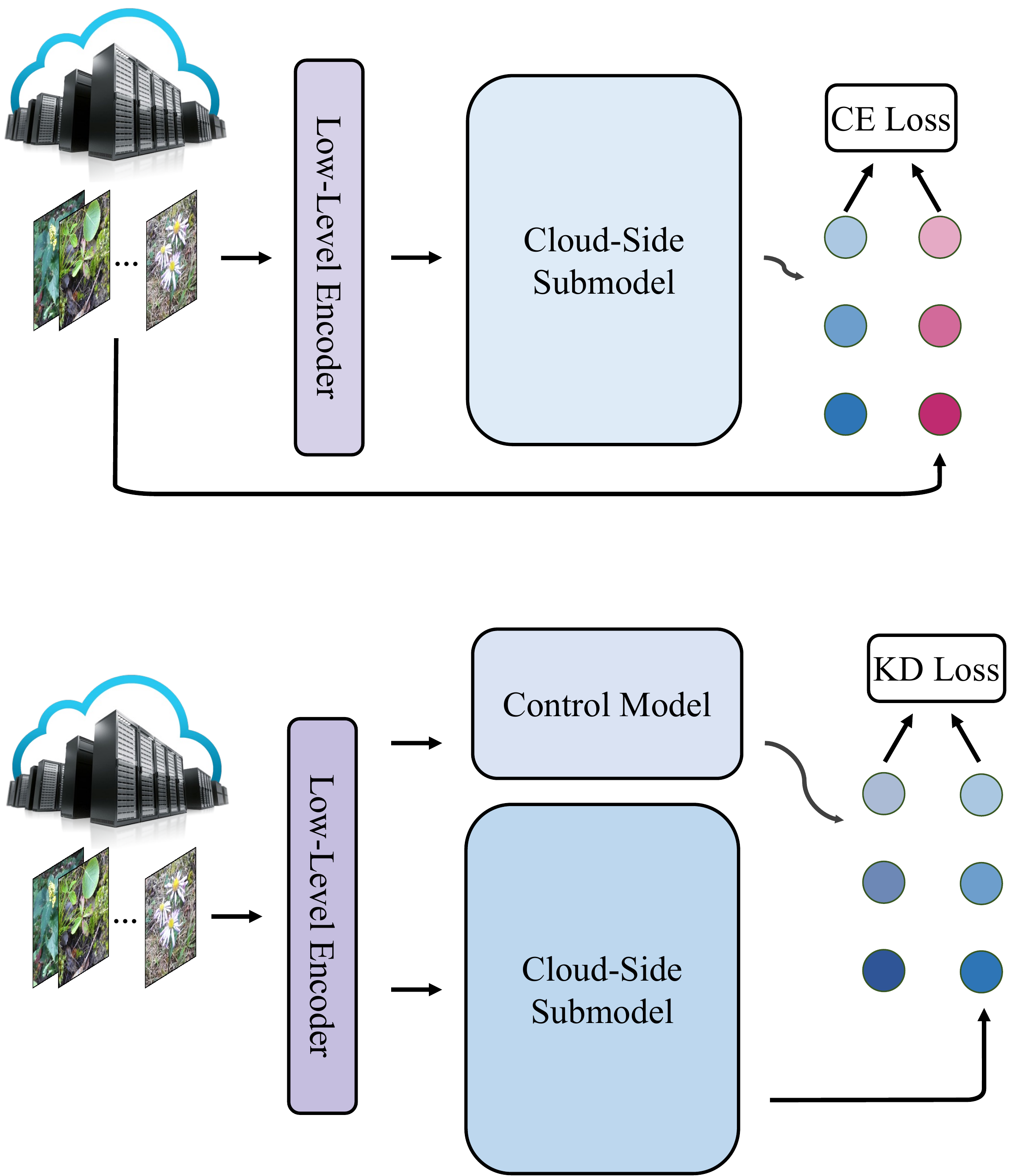}
}
\subfigure[Device-Cloud Collaborative Training]{
\includegraphics[width=0.28\linewidth]{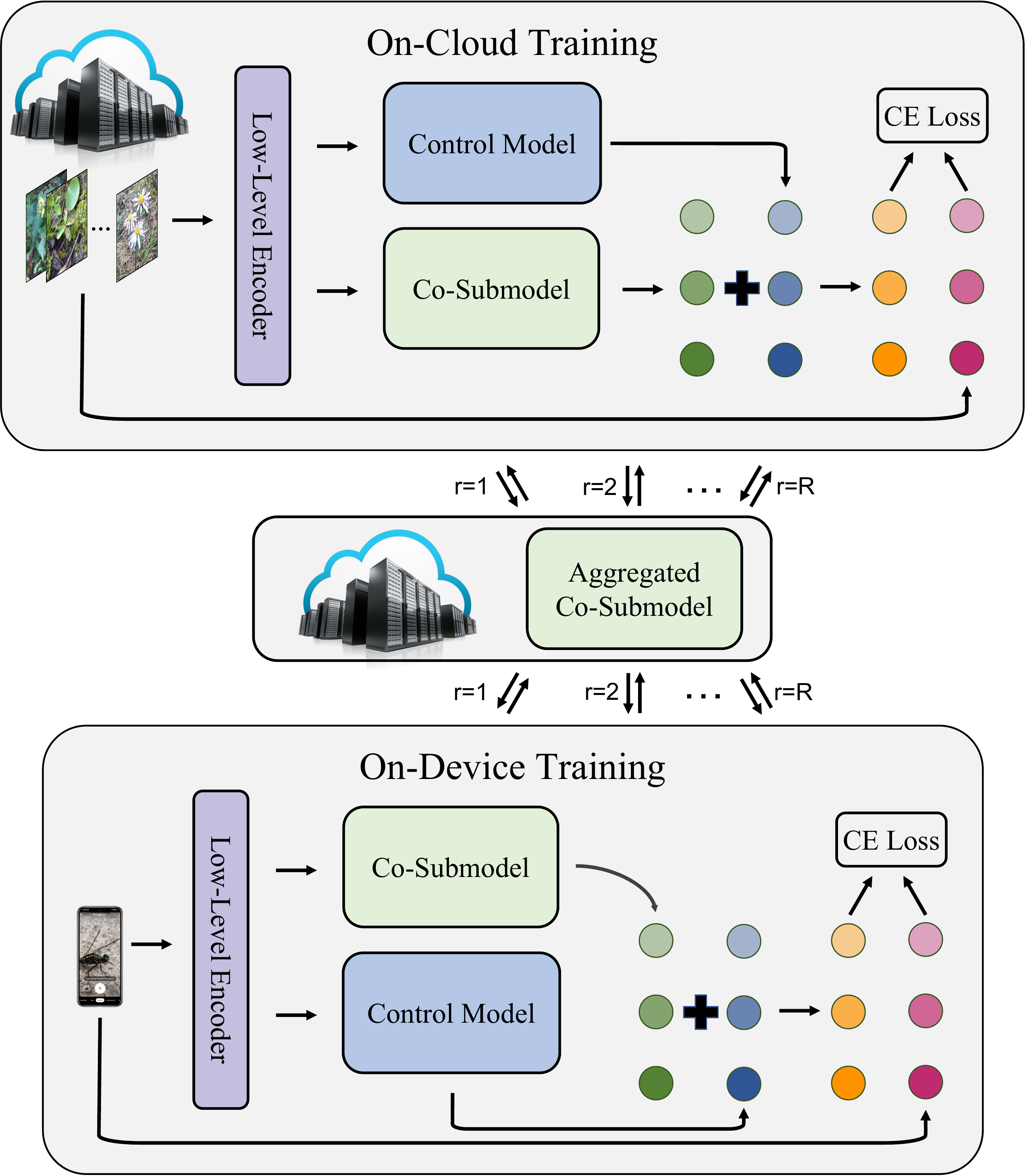}
}
\subfigure[Classifier Finetuning]{
\includegraphics[width=0.35\linewidth]{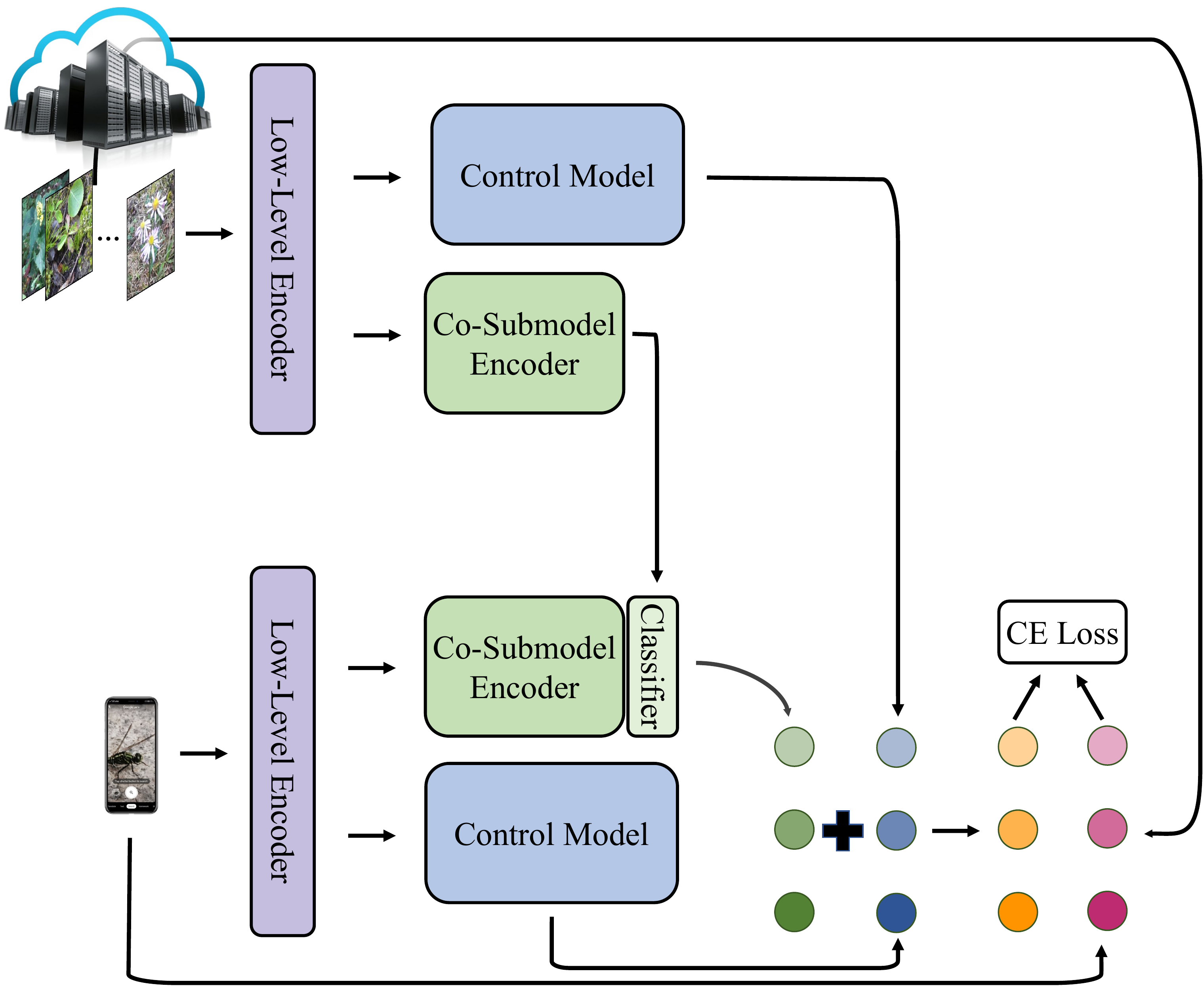}
}
\caption{Workflow of DC-CCL: (1) the cloud trains the shared encoder and the cloud-side submodel; (2) the cloud trains the light-weight control model to mimic the large cloud-side submodel with knowledge distillation; (3) the cloud and the mobile device collaboratively train the co-submodel under the correction of the control model; and (4) the classifier of the co-submodel is finetuned to mitigate device-cloud sample skewness.}
\label{DC-CCL_workflow}
\end{figure*}


In this section, we introduce the design details of DC-CCL. We depict the whole workflow in Figure \ref{DC-CCL_workflow}. From device-cloud submodel structures, a low-level feature encoder is shared at the bottom of the large cloud-side submodel, and the light-weight control model, and the small device-cloud co-submodel (Section \ref{sms}). The cloud first trains the shared encoder and the cloud-side submodel over its samples, and then trains the control model to mimic the cloud-side submodel through knowledge distillation (Section \ref{TCM}). Then, the mobile device and the cloud collaboratively train the co-submodel under the correction of the control model in a data parallelism way, and finetune the classifier to mitigate device-cloud sample skewness (Section \ref{dcdt}). We finally introduce how to support that the cloud-side submodel takes a pre-trained model, while the co-submodel and the control model take different backbone network architectures from the pre-trained model (Section \ref{sec:pretrain}).

\subsection{Device-Cloud Submodel Structures}\label{sms}

A base vision model, typically a CNN, from the bottom to the top, is normally stacked by a low-level feature encoder, a high-level feature encoder, and a classifier. The low-level encoder comprises a few shallow layers, whereas the high-level encoder contains many dense layers and dominates the size of the base model. In DC-CCL, the low-level feature encoder of the base model is shared at the bottom of the cloud-side submodel and the device-cloud co-submodel to extract basic and common features. In contrast, the high-level feature encoder and the classifier of two submodels are completely separated. More specifically, with the same low-level features as input, the feature mappings as well as the processes of activation forward and gradient backward are functionally decoupled, and the final outputs (i.e., the logits of the classifier) are aligned and aggregated. The architecture design of the high-level encoders in two submodels can follow the same backbone as the base model's high-level encoder, but take a varying number of filters in each network layer to scale up or scale down the parameter size. We let $\alpha_{cl}$ (resp., $\alpha_{co}$) denote the ratio between the number of filters in the cloud-side submodel (resp., the device-cloud co-submodel) and the number of filters in the base model. If $\alpha_{cl} + \alpha_{co} = 1$, the construction process of two submodels can also be viewed from vertically splitting the base model and cutting off the neural connections, as depicted in Figure \ref{iso_model_arc}. Of course, DC-CCL allows adjusting $\alpha_{cl}$ and $\alpha_{co}$ to adapt to the resource richness and limitation of the cloud and the mobile device. For example, if $\alpha_{co} = \frac{1}{8}$, the numbers of input and output filters are both reduced to their $\alpha_{co} = \frac{1}{8}$, and the device-cloud co-submodel size is reduced to roughly $\alpha_{co}^2=\frac{1}{64}$ of the base model size.

\subsection{Cloud-Based Training}\label{TCM}

The cloud first trains the shared low-level feature encoder and the cloud-side submodel for several epochs. Then, the cloud trains a light-weight control model with knowledge distillation to mimic the cloud-side submodel. Isomorphic to the submodels, the control model shares the low-level encoder at the bottom, takes the same backbone as the base model, but sets a small number of filters. Additionally, the knowledge distillation loss over the cloud-side dataset $D_c$ is:
\begin{equation*}
\begin{aligned}
    L_{KD} \overset{\triangle}{=} \frac{1}{|D_c|}\sum_{(x_i,y_i)\in D_c} l_{kd}\left(cloud\_output_i, control\_output_i\right),\\
\end{aligned}
\end{equation*}
where $l_{kd}(\cdot, \cdot)$ takes mean squared error (MSE), $cloud\_output_i$ and $control\_output_i$ denote the outputs of the cloud-side submodel and the control model, respectively. After several epochs of knowledge distillation, the output of the control model can approximate that of the cloud-side submodel.  

\subsection{Device-Cloud Collaborative Training}\label{dcdt}

Section \ref{ftim} has demonstrated the feasibility of training the device-cloud co-submodel under the correction of the control model over the global dataset. We now adapt the centralized training of the co-submodel in a practical distributed way. We adopt the classical data parallelism framework, where the cloud and the mobile device keep their samples locally and train the co-submodel over their respective samples. During local training, besides freezing the control model, the shared encoder, the output of which functions as the input of the control model, should also be frozen to ensure the effectiveness of the control model in mimicking the cloud-side submodel. In addition, the cloud-side or the device-side training loss is derived based on the sum of the logits of the control model and the co-submodel, formally defined as
\begin{equation*}
    L \overset{\triangle}{=} \frac{1}{|B|}\sum_{(x_i,y_i)\in B} l_{CE}\left(control\_output_i + co\_output_i, y_i\right), 
\end{equation*}
where $(x_i,y_i)$ denotes a sample from the cloud-side or the device-side data batch $B$, $l_{CE}(\cdot,\cdot)$ takes the cross entropy (CE) loss function, and $control\_output_i$ and  $co\_output_i$ denote the outputs of the control model and the co-submodel, respectively. After multiple iterations of local training, the cloud, as a parameter server, aggregates its updated co-submodel with the updated co-submodel from the mobile device and generates a new co-submodel, which works as the starting point of the next round collaborative training. 

If the distributions of the samples on the cloud and on the mobile device significantly differ from each other (e.g., involving different classes of images), and are long-tailed from the perspective of the global distribution (i.e., missing the samples on the other side), the local updates of device-cloud co-submodel will be skew, degrading the performance of collaborative training. Existing work \cite{lt1, lt2} ever studied cloud-based training over long-tailed distribution and proposed to adjust the classifier. In our device-cloud context, after standard collaborative training, we propose to finetune the co-submodel's classifier over the global dataset, as required. Finetuning the classifier mainly needs the outputs of the co-submodel's high-level encoder over the cloud-side and the device-side samples. Specifically, the size of the high-level features is much smaller than the size of the raw sample. For example, a 224$\times$224 image with RGB channels occupies 150,528 bytes, while the corresponding VGG16's high-level features take only 512 bytes, which is 0.3\% of the raw image size. 
Therefore, if the high-level features of the device-side samples are sensitive and cannot be uploaded to the cloud, classifier finetuning should be performed on the mobile device, and the high-level features of the cloud-side insensitive samples are efficient to be downloaded. Otherwise, the finetuning can be performed on the cloud.

\subsection{Supporting Pre-Trained Models}\label{sec:pretrain}


\begin{figure}[!t]
\centering
\includegraphics[width=0.77\columnwidth]{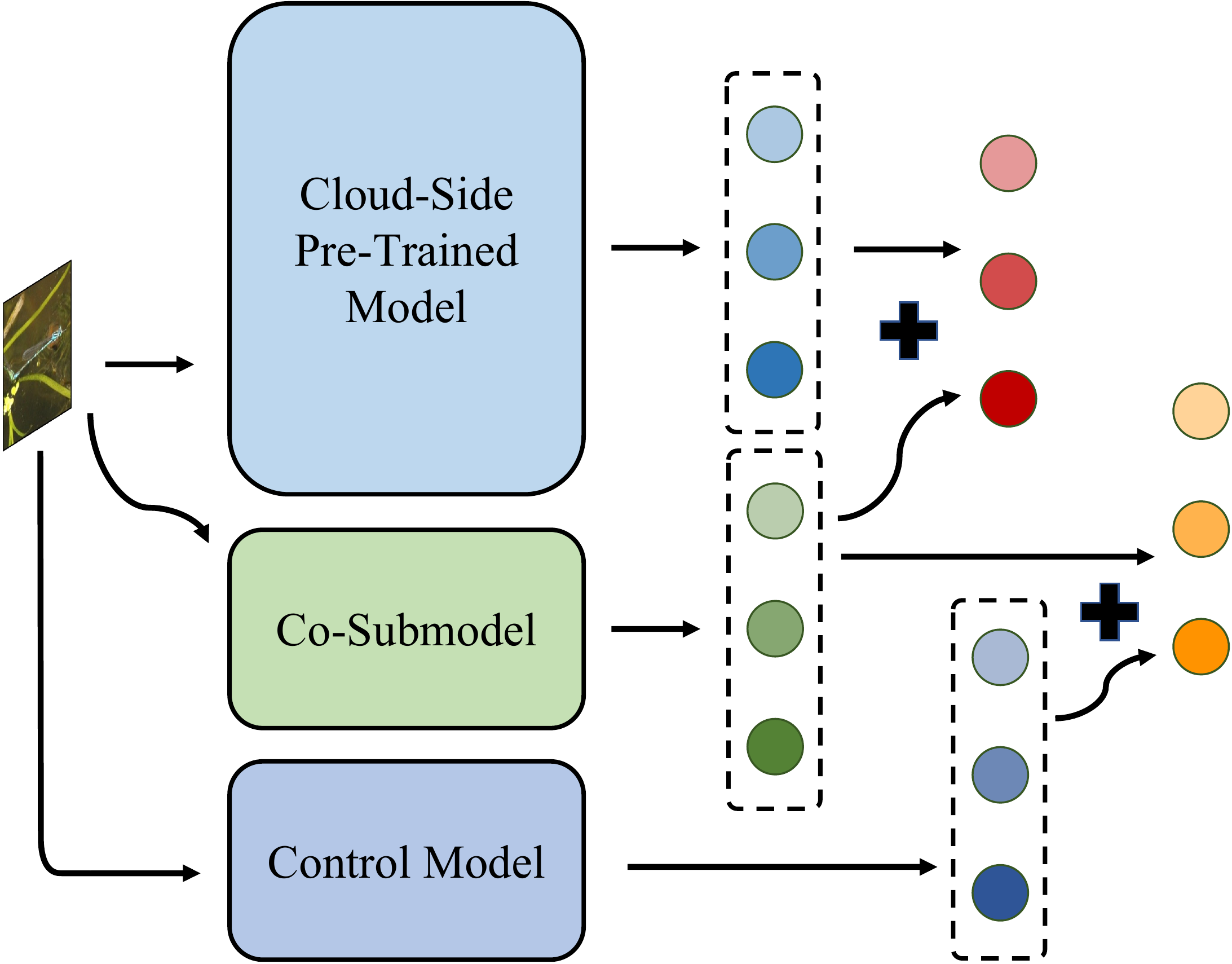}
\caption{DC-CCL with cloud-side pre-trained model, while the device-cloud co-submodel and the control model take different backbone architectures.}
\label{pre-str}
\end{figure}

The design of DC-CCL above starts from a base model with an existing backbone network architecture, decouples it into the shared low-level encoder, the cloud-side submodel, and the device-cloud co-submodel, and trains them from scratch. We further consider a common case that the cloud-side submodel takes a large pre-trained model. If the device-cloud co-submodel and the control model adopt the same backbone network architecture with the cloud-side pre-trained model, then our design still applies. If the co-submodel and the control model take different backbone architectures from the cloud-side pre-trained model (e.g., MobileNet vs. EfficientNet), then the key difference is that as shown in Figure~\ref{pre-str}, no low-level feature encoder is shared at the bottom any more.   

%% file: 04Evaluation.tex
\section{Evaluation}\label{sec_eval}

In this section, we first introduce evaluation setup (Section \ref{sec:eval:setup}); then compare with ideal and practical baselines (Sections \ref{sec:eval:large} and \ref{sec:eval:small}); further show ablation study (Section \ref{sec:eval:abl}); and finally reveal device-side model flexibility (Section \ref{sec:eval:device}).   

\begin{table*}[!t]
    \caption{DC-CCL vs. Baselines with large model. The class ratio denotes the ratio between the number of cloud-side sample classes and that of the device-side classes. $\Delta$ denotes the accuracy improvement of DC-CCL over Cloud-B.}
    \vspace{-0.3em}
    \label{acc_cmp_ideal}
    \centering
    \resizebox{0.93\textwidth}{!}{
    \begin{tabular}{cccccccccc}
    \toprule
\multirow{2}{*}{Dataset}     & \multirow{2}{*}{Central-D} & \multirow{2}{*}{Central-B} & \multicolumn{2}{c}{Model Size (MB)}
& \multirow{2}{*}{Class Ratio} & \multicolumn{1}{c}{\multirow{2}{*}{Distr-D}} & \multirow{2}{*}{Cloud-B} & \multirow{2}{*}{DC-CCL} & \multirow{2}{*}{$\Delta$} \\
\cmidrule{4-5}
 &  &  & Cloud-Side & Device-Side & & & &  \\
    \midrule
\multirow{4}{*}{CIFAR10} & \multirow{4}{*}{84.38\%} & \multirow{4}{*}{86.01\%} & \multirow{4}{*}{98.42} & \multirow{4}{*}{4.27} & 1/9 & \multicolumn{1}{c|}{79.77\%} & 76.63\% & 80.60\% & +3.97\%\\
            & & & &    & 2/8 & \multicolumn{1}{c|}{79.42\%} & 67.95\% & 80.32\% & +12.37\% \\
            & & & &    & 3/7 & \multicolumn{1}{c|}{76.38\%} & 60.10\% & 77.40\% & +17.30\% \\
            & & & &    & 4/6 & \multicolumn{1}{c|}{77.10\%} & 54.33\% & 77.31\% & +22.98\% \\
    \midrule
\multirow{4}{*}{CIFAR100} & \multirow{4}{*}{67.21\%}  & \multirow{4}{*}{67.52\%}  & \multirow{4}{*}{32.83} & \multirow{4}{*}{1.43} & 10/90 & \multicolumn{1}{c|}{66.22\%} & 61.75\% & 65.27\% & +3.52\% \\
            & & & &   & 20/80 & \multicolumn{1}{c|}{57.98\%} & 56.54\% & 62.74\% & +6.20\% \\
            & & & &   & 30/70 & \multicolumn{1}{c|}{60.11\%} & 49.65\% & 61.65\% & +12.00\% \\
            & & & &   & 40/60 & \multicolumn{1}{c|}{61.34\%} & 43.80\% & 58.55\% & +14.75\% \\
    \midrule
\multirow{4}{*}{StanfordCars} & \multirow{4}{*}{66.12\%}   & \multirow{4}{*}{71.01\%}   & \multirow{4}{*}{12.90} & \multirow{4}{*}{3.11}  & 20/176 & \multicolumn{1}{c|}{66.71\%} &  59.05\% & 63.20\% & +4.15\% \\
            & & & &   & 30/166 & \multicolumn{1}{c|}{64.27\%} & 55.15\% & 59.61\% & +4.46\% \\
            & & & &   & 40/156 & \multicolumn{1}{c|}{62.28\%} & 52.53\% & 57.05\% & +4.52\% \\
            & & & &   & 50/146 & \multicolumn{1}{c|}{61.53\%} & 47.68\% & 55.20\% & +7.52\% \\
    \midrule
\multirow{3}{*}{UCF101} & \multirow{3}{*}{86.56\%}  & \multirow{3}{*}{87.45\%}  & \multirow{3}{*}{229.28} & \multirow{3}{*}{9.91} & 10/101 & \multicolumn{1}{c|}{85.63\%} & 79.04\% & 83.78\% & +4.74\% \\
            & & & &   & 20/91 & \multicolumn{1}{c|}{83.67\%} & 71.08\% & 83.89\% & +12.81\% \\
            & & & &   & 30/81 & \multicolumn{1}{c|}{83.82\%} & 61.90\% & 81.23\% & +19.33\% \\
    \midrule
ImageNet-1K \&   & \multirow{2}{*}{72.93\%} & \multirow{2}{*}{69.42\%} & \multirow{2}{*}{75.12} &  \multirow{2}{*}{20.93} & \multirow{2}{*}{196/1000} & \multicolumn{1}{c|}{\multirow{2}{*}{69.38\%}} &  \multirow{2}{*}{60.11\%} & \multirow{2}{*}{69.32\%} & \multirow{2}{*}{+9.21\%} \\
StanfordCars & & & & & & \multicolumn{1}{c|}{} & &\\
    \bottomrule
    \end{tabular}
    }
\end{table*}

\subsection{Experimental Setup}\label{sec:eval:setup}


\quad\textbf{Datasets.} We use CIFAR10~\cite{cifar}, CIFAR100~\cite{cifar}, StanfordCars~\cite{cars_data}, ImageNet-1K~\cite{imagenet}, and UCF101~\cite{ucf101}. CIFAR10 and CIFAR100 comprise 10 and 100 classes, respectively, 50,000 training images and 10,000 testing images. StanfordCars consists of 196 classes of car images, 8,144 training images and 8,041 testing images. ImageNet-1K involves 1,000 classes and comprises 1,281,167 training images and 50,000 testing images. UCF101 contains 13,320 videos of 101 action classes, 80\% for training and 20\% for testing.

\textbf{Device-Cloud Sample Partition.} We divide the full training set of CIFAR10, CIFAR100, StanfordCars, or UCF101 into two subsets according to labels, one for the mobile device and the other for the cloud. We also vary the number of device-side sample classes, and the number of cloud-side sample classes varies as well. For CIFAR10 and CIFAR100, we augment the device-side training set by randomly introducing a few cloud-side samples, the size of which is less than 10\% of the size of the original device-side training set. In the inference phase, the accuracy is computed over each dataset's full test set. We also leverage ImageNet-1K and StanfordCars to evaluate the case of pre-trained model, where the cloud keeps the training set of ImageNet-1K, and the mobile device holds the training set of StanfordCars, while the inference accuracy is evaluated over the union of the test sets of two datasets. To mitigate the long-tail distribution due to the difference between ImageNet-1K and StanfordCars, the cloud randomly chooses only $8,144 / 196 \times 1,000 \approx 40,000$ training images from ImageNet-1K, achieving the best performance.

\textbf{Base Models and Device-Cloud (Sub)models.} For the image classification task over CIFAR10, CIFAR100, and StanfordCars, as well as the action recognition task over UCF101, we take VGG16~\cite{vgg}, ResNet18~\cite{resnet}, EfficientNet-B0~\cite{efficientnet}, and C3D~\cite{c3d} as the base model, respectively. We choose the first layer as the shared low-level encoder and perform vertical splitting for the upper layers. In particular, we set the number filters in the cloud-side submodel and the device-cloud co-submodel $\alpha_{cl}=\frac{7}{8}$ and $\alpha_{co}=\frac{1}{8}$ of that in the base model, respectively, by default. We also vary $\alpha_{co}$ to scale up and scale down the co-submodel. For the cloud-side pre-trained model over ImageNet-1K, we take EfficientNet-B4~\cite{efficientnet}, while adopting a pre-trained MobileNetV3-small~\cite{mobilenetv3-small} as the device-cloud co-submodel. In each evaluation, the control model takes the same structure as the device-cloud co-submodel. We list the size of the cloud-side model (i.e., the shared encoder, the cloud-side submodel, and the co-submodel) and the size of the device-side model (i.e., the shared encoder, the control model, and the co-submodel) in  Table~\ref{acc_cmp_ideal}.

\textbf{Implementation Details.} We implement DC-CCL with {PyTorch 3.7}. We take adaptive moment estimation (Adam) as the optimizer of the knowledge distillation for the control model and take stochastic gradient descent (SGD) as the optimizer of all the other training phases. For device-cloud collaborative training, we set the total number of communication rounds to 15, 20, 25, 25, and 10 for CIFAR10, CIFAR100, StanfordCars, UCF101, and ImageNet-1K \& StanfordCars. More details on the learning rates, the number of epochs for training cloud-side submodel, knowledge distillation, and classifier finetuning, are deferred to {Table \ref{localepochs}} in Appendix \ref{exp_detail}.

\begin{figure*}[!t]
\begin{minipage}{0.62\linewidth}
\centering 
\begin{minipage}{0.8\textwidth} 
\centering 
\includegraphics[width=\textwidth]{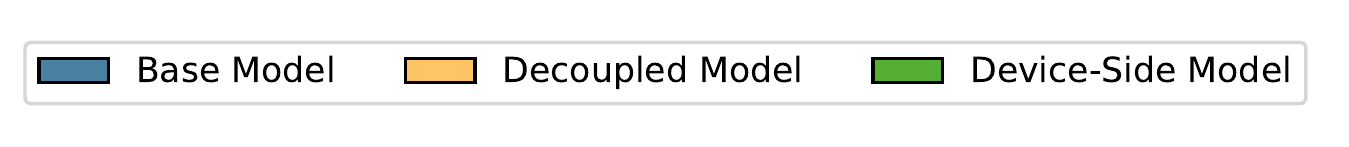}
\end{minipage}
\begin{minipage}{\textwidth} 
\vspace{-0.6em}
\centering
\subfigure[Parameters Size]{
\includegraphics[width=0.47\textwidth]{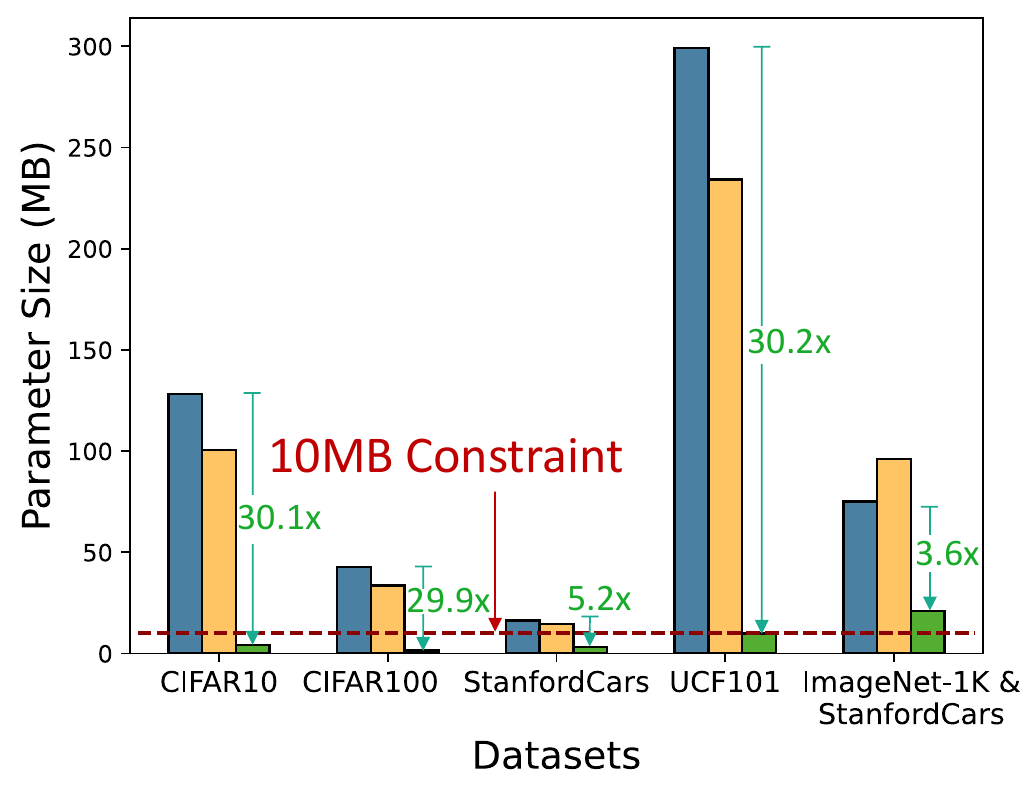}
}
\subfigure[FLOPs]{
\includegraphics[width=0.47\textwidth]{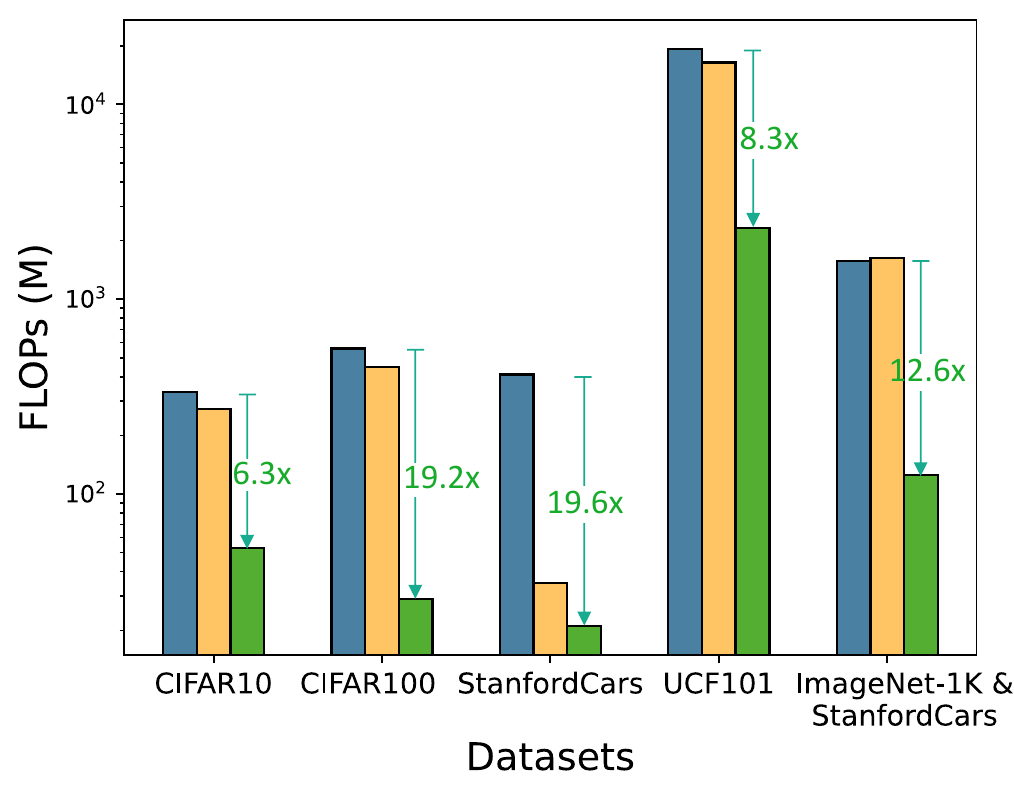}
}
\end{minipage}
\vspace{-0.8em}
\caption{The sizes and FLOPs of the base model, the decoupled model, and the device-side model.} \label{model-efficiency}
\end{minipage}
\hfill
\begin{minipage}{0.35\linewidth}
\vspace{0.6em}
\begin{minipage}{\textwidth} 
\centering 
\includegraphics[width=0.93\textwidth]{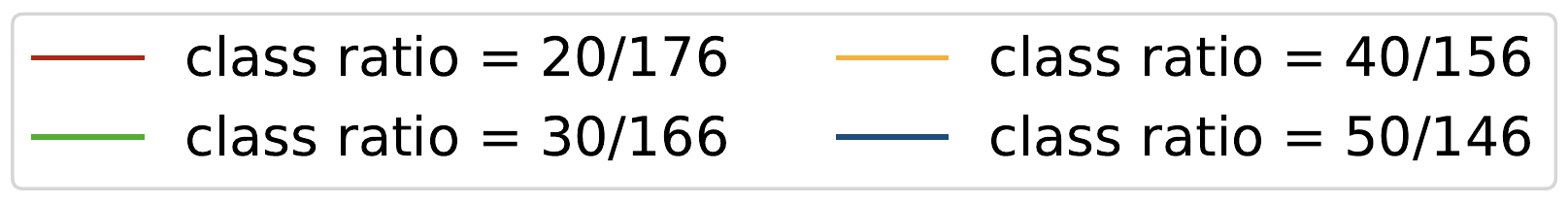}
\end{minipage}
\begin{minipage}{\textwidth} 
\centering 
\includegraphics[width=0.95\textwidth]{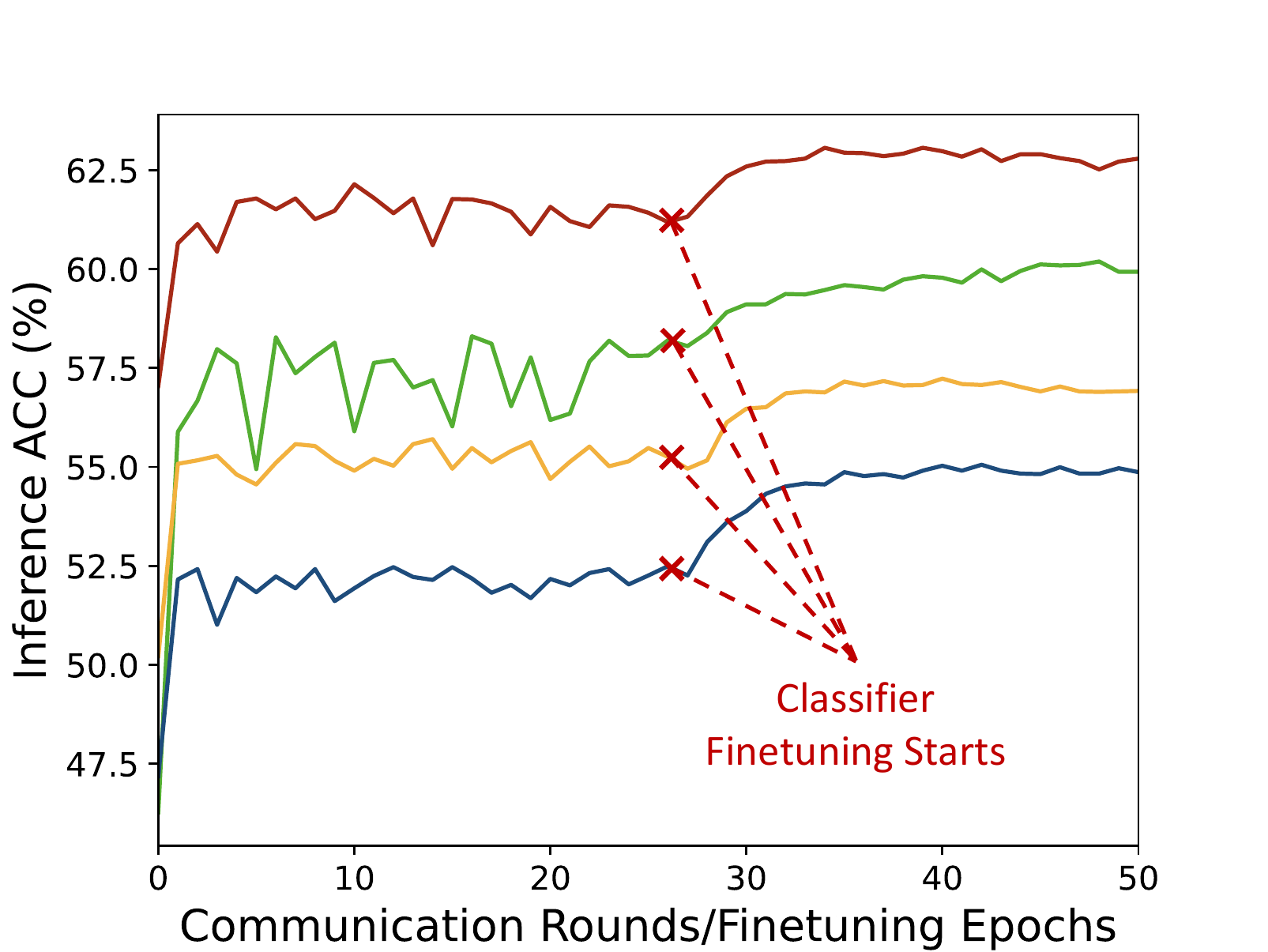}
\vspace{0.1em}
\caption{Device-cloud collaborative learning process over StanfordCars.}
\label{acc_curve}
\end{minipage}
\end{minipage}
\end{figure*}

\subsection{DC-CCL vs. Baselines with Large Model}\label{sec:eval:large}

We first validate the necessity of exploiting both cloud-side and device-side samples. We introduce two baselines:
\begin{itemize}
    \item {\bf Central-B}, which ideally puts aside that the device-side samples cannot be uploaded to the cloud and trains the base model over the full training set;
    \item {\bf Cloud-B}, which trains the base model only over the cloud-side training set.
\end{itemize}
We further consider exploiting the full training set under the practical settings that the vision samples are distributed on the mobile device and the cloud, and the resource-constraint mobile device cannot hold a large model. We introduce another two baselines:
\begin{itemize}
    \item {\bf Central-D}, which ideally puts aside that the device-side samples cannot be uploaded to the cloud and trains the decoupled model over the full training set;  
    \item {\bf Distr-D}, which ideally puts aside that the decoupled model is unaffordable to be offloaded to the mobile device, lets the mobile device and the cloud train the decoupled model in a data parallelism way, and also applies classifier finetuning as DC-CCL.  
\end{itemize}
We report in Table \ref{acc_cmp_ideal} the inference accuracy of the four baselines and the proposed DC-CCL as well as the number of sample classes and the model size on the sides of the mobile device and the cloud.

By comparing Central-B with Cloud-B, we can see that as the size of device-side samples becomes larger, the advantage of Central-B over Cloud-B is more remarkable. For example, for StanfordCars and UCF101, when the number of device-side sample classes is roughly 1/3 of the number of cloud-side sample classes, the inference accuracy of Central-B is at least 23\% higher than that of Cloud-B. These results demonstrate that it is quite necessary to optimize the model over both the cloud-side and the device-side samples. 

By comparing Distr-D with Central-D, we can see that Distr-D underperforms Central-D, especially over CIFAR10. In addition, when the sample classes are more balanced distributed on the mobile device and the cloud, the performance of Distr-D generally becomes worse. Such performance degradation is mainly due to the practical non-independent and identically distributed (non-iid) feature of the device-side and the cloud-side samples.

\begin{table*}[!t]
    \caption{DC-CCL vs. Baselines with small model vs. DC-CCL without the control model or the classifier finetuning. $\Delta_1$ and $\Delta_2$ denote the accuracy improvements of DC-CCL over Distr-S and Incr-S, respectively.}
    \label{acc_cmp_light}
    \centering
    \resizebox{1.95\columnwidth}{!}{
    \begin{tabular}{ccccccccc}
    \toprule
Dataset     & 
Class Ratio & Distr-S & Incr-S & DC-CCL & $\Delta_1$ & $\Delta_2$ & - Control Model & - Finetuning \\
    \midrule
\multirow{4}{*}{CIFAR10}  
& 1/9   & 75.26\%      & 68.99\%    & 80.60\%  & +5.34\% & \multicolumn{1}{c|}{+11.61\%} & -1.89\%     & -17.30\% \\
            & 2/8   & 70.28\%      & 66.35\%   & 80.32\% & +10.04\% & \multicolumn{1}{c|}{+13.97\%} & -10.08\%    & -9.19\% \\
            & 3/7   & 64.63\%      & 68.18\%   & 77.40\% & +12.77\% & \multicolumn{1}{c|}{+9.22\%} & -13.29\%    & -16.06\%  \\
            & 4/6   & 61.69\%      & 66.62\%   & 77.31\% & +15.62\% & \multicolumn{1}{c|}{+10.69\%} & -19.44\%    & -14.00\% \\
    \midrule
\multirow{4}{*}{CIFAR100}  
& 10/90  & 52.36\%      & 44.59\%   & 65.27\% & +12.91\% & \multicolumn{1}{c|}{+20.61\%} & -2.71\%     & -4.42\%\\
            & 20/80  & 51.64\%      & 45.59\%   & 62.74\% & +11.10\% & \multicolumn{1}{c|}{+17.15\%} & -5.37\%     & -6.83\%  \\
            & 30/70  & 48.69\%      & 46.81\%   & 61.65\% & +12.96\% & \multicolumn{1}{c|}{+14.84\%} & -11.90\%    & -6.21\% \\
            & 40/60  & 46.15\%      & 46.73\%   & 58.55\% & +12.40\% & \multicolumn{1}{c|}{+11.82\%} & -14.39\%    & -4.76\% \\
    \midrule
\multirow{4}{*}{StanfordCars}  
& 20/176  & 54.87\% & 39.26\%  & 63.20\% & +8.33\% & \multicolumn{1}{c|}{+23.94\%} & -1.76\%     & -1.42\% \\
            & 30/166  & 54.10\% & 34.78\%   & 59.61\% & +5.51\% & \multicolumn{1}{c|}{+24.83\%} & -2.89\%     & -1.31\% \\
            & 40/156  & 52.02\% & 34.18\%   & 57.05\% & +5.03\% & \multicolumn{1}{c|}{+22.87\%} & -1.47\%     & -1.35\% \\
            & 50/146  & 48.13\% & 32.94\%   & 55.20\% & +7.07\% & \multicolumn{1}{c|}{+22.26\%} & -2.56\%     & -2.73\% \\
    \midrule
\multirow{3}{*}{UCF101} 
& 10/91 & 78.60\% & 63.94\%  & {83.78\%}  & +5.18\% & \multicolumn{1}{c|}{+19.84\%} & -4.48\%     & -6.62\% \\
            & 20/81  & 76.34\% & 65.72\% & 83.89\% & +7.55\% & \multicolumn{1}{c|}{+18.17\%} & -13.88\%     & -25.87\% \\
            & 30/71  & 75.82\% & 69.53\% & 81.23\% & +5.41\% & \multicolumn{1}{c|}{+11.70\%} & -20.25\%     & -27.95\% \\
    \midrule
ImageNet-1K \& 
& \multirow{2}{*}{196/1000} & \multirow{2}{*}{39.74\%} & \multirow{2}{*}{28.00\%}  & \multirow{2}{*}{69.32\%} & \multirow{2}{*}{+29.58\%} & \multicolumn{1}{c|}{\multirow{2}{*}{+41.32\%}} & \multirow{2}{*}{-6.76\%} & \multirow{2}{*}{-3.60\%} \\
StanfordCars & & & & & & \multicolumn{1}{c|}{} \\ 
    \bottomrule
    \end{tabular}
    }
\end{table*}

By comparing DC-CCL with Distr-D, we can see that their inference accuracy are close. Averagely, DC-CCL decreases the inference accuracy only by 1.16\%, compared with Distr-D. The slight accuracy drop comes from the practical infeasibility of deploying the large model on the mobile device. We also compare the size and the FLOPs (floating point operations) of the models deployed on the mobile device in Distr-D and DC-CCL. From Figure \ref{model-efficiency} and Table \ref{acc_cmp_ideal}, we can observe that (1) in DC-CCL, the sizes of the device-side models for CIFAR10, CIFAR100, StanfordCars, and UCF101 are less than 10MB. For ImageNet-1K \& StanfordCars, the mobile device take two MobileNetV3-small as the co-submodel and the control model, the total size of which is 20.93MB; and (2) compared with the light-weight device-side models in DC-CCL, the decoupled models on the mobile device in Distr-D are roughly $24\times$, $23\times$, $5\times$, $24\times$, and $5\times$ larger from parameter size, and $5\times$, $16\times$, $2\times$, $7\times$, and $13\times$ more time-consuming from FLOPs, for CIFAR10, CIFAR100, StanfordCars, UCF101, and ImageNet-1K \& StanfordCars, respectively. We further compare the communication overhead. DC-CCL needs to exchange the device-cloud co-submodel (update) in each communication round, while  Distr-D exchanges the decoupled model (update). In addition, the number of communication rounds required by DC-CCL and Distr-D are consistent. Therefore, compared with DC-CCL, Distr-D is roughly $48\times$, $46\times$, $10\times$, $48\times$, and $10\times$ more bandwidth-consuming. We also plot in Figure \ref{acc_curve} the device-cloud collaborative training process of DC-CCL over StanfordCars, which requires the most communication rounds among all the datasets. We can see that the number of communication rounds is no more than 25. The results above demonstrate that DC-CCL approaches the ideal baseline Distr-D from accuracy, enables device-cloud collaborative learning of the large vision models, and sharply reduces computation and communication overhead.

By comparing DC-CCL with Cloud-B, DC-CCL increases the inference accuracy on average (i.e., over different settings of the class ratio) by 14.16\%, 9.12\%, 5.16\%, 12.29\%, and 9.21\%, for CIFAR10, CIFAR100, StanfordCars, UCF101, and ImageNet-1K \& StanfordCars, respectively. These results reveal the benefit of leveraging device-side samples using our design under the practical setting of no raw sample up-link and no large model down-link.

\subsection{DC-CCL vs. Baselines with Small Model}\label{sec:eval:small}

Considering the fact that the mobile device cannot hold a large model, a practical and trivial idea is using an affordable small model, typically, the shared encoder plus the device-cloud co-submodel in DC-CCL. We introduce two baselines:
\begin{itemize}
    \item {\bf Distr-S}, which differs from Distr-D only in that the mobile device and the cloud leverage the shared encoder plus the co-submodel rather than the large decoupled model; 
    \item {\bf Incr-S}, which trains the shared encoder plus the co-submodel in an incremental learning way, first over the cloud-side samples and then over the device-side samples, and finally applies classifier finetuning.
\end{itemize}
We report in Table \ref{acc_cmp_light} the accuracy of the baselines and DC-CCL. From Table \ref{acc_cmp_light}, we can observe that compared with Distr-S, DC-CCL improves the average accuracy by {10.94\%, 12.34\%, 6.49\%, 6.05\%, and 29.58\%}; and compared with Incr-S, DC-CCL improves the average accuracy by {11.37\%, 16.12\%, 23.48\%, 16.58\%, and 41.32\%}, for CIFAR10, CIFAR100, StanfordCars, UCF101, ImageNet-1K \& StanfordCars, respectively. We can draw that it is necessary to maintain a high model capacity in device-cloud collaborative learning.

\subsection{Ablation Study}\label{sec:eval:abl}

To validate the necessity of each ingredient in DC-CCL, we first remove the control model. Specifically, after training the shared low-level encoder and the cloud-side submodel over the cloud-side training set, these parameters are frozen, and only the device-cloud co-submodel is trained over the full training set. We also remove the classifier finetuning step from DC-CCL to reveal its necessity. As shown in Table \ref{acc_cmp_light}, without the control model, the average accuracy drops by {11.18\%, 8.59\%, 2.17\%, 12.87\%, and 6.76\%}; and without the classifier finetuning, the average accuracy drops by {14.14\%, 5.56\%, 1.70\%, 20.15\%, and 3.60\%}, for CIFAR10, CIFAR100, StanfordCars, UCF101, ImageNet-1K \& StanfordCars, respectively. The  results above validate the compactness of DC-CCL.

\subsection{Scaling of Device-Side Model Size}\label{sec:eval:device}

\begin{figure}[!t]
    \centering
    \subfigure[CIFAR10]{
    \includegraphics[width=0.472\columnwidth]{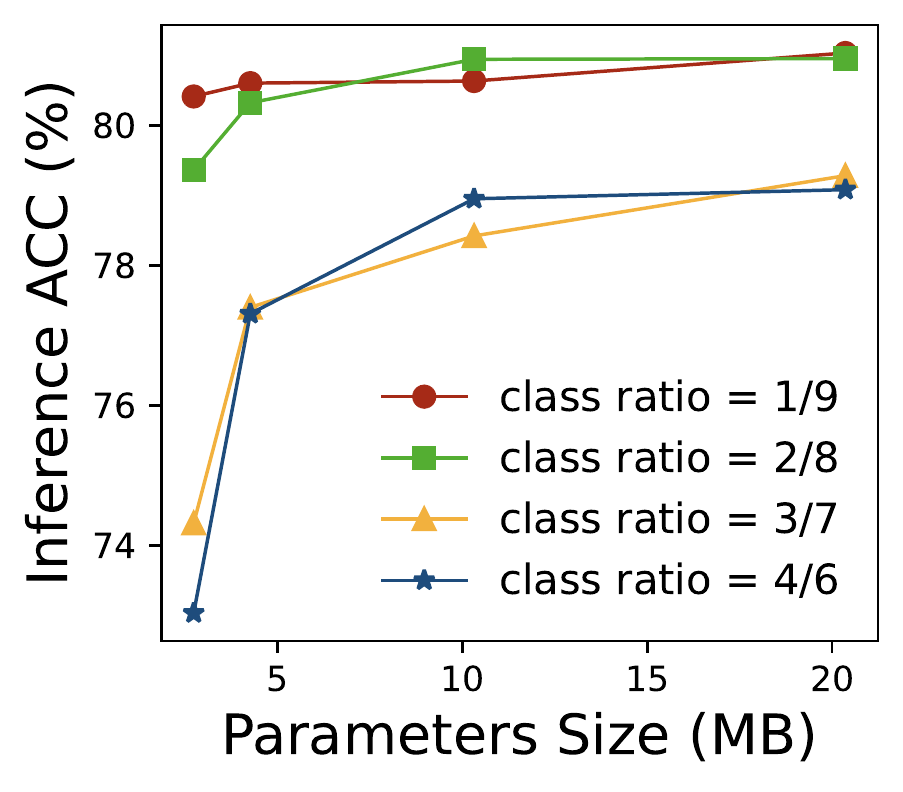}
    }
    \subfigure[CIFAR100]{
    \includegraphics[width=0.472\columnwidth]{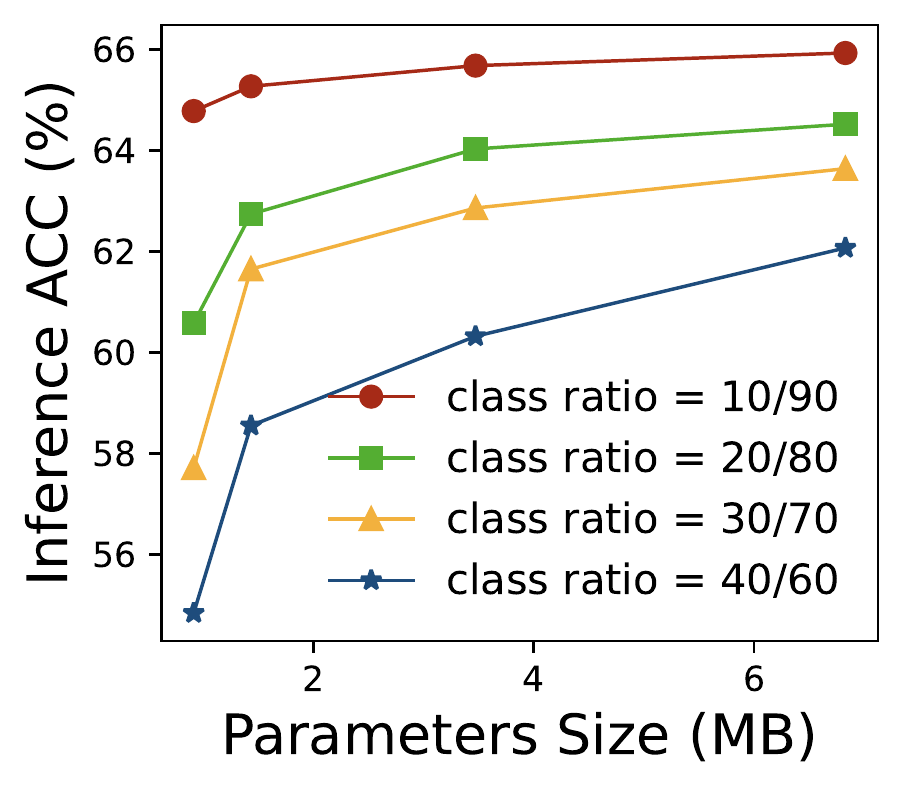}
    }
    \caption{Accuracy of DC-CCL with the varying device-side model size over CIFAR10 and CIFAR100.}\label{device-modelsize}
\end{figure}

We finally evaluate how scaling the size of the device-side model (i.e., the shared encoder, the device-cloud co-submodel, and the control model) will affect the performance of DC-CCL. In particular, we keep the shared encoder and the control model by default and vary the number filters in the device-cloud co-submodel, accounting for $\alpha_{co}$ of the number of filters in the base model. We let $\alpha_{co}$ decrease from the default setting $\frac{1}{8}$ to $\frac{1}{16}$ and increase to $\frac{1}{8}$, $\frac{1}{4}$, and $\frac{3}{8}$, such that the decoupled model is no larger than the base model (i.e., $\alpha_{co}^2 + \alpha_{cl}^2 \leq \frac{3}{8}^2 + \frac{7}{8}^2 < 1$). We plot in Figure \ref{device-modelsize} the accuracy of DC-CCL and the device-side model size. One key observation is that compared with the default setting $\alpha_{co} = \frac{1}{8}$, DC-CCL with $\alpha_{co}=\frac{1}{16}$ decreases the device-side model size by {42\% and 36\%}, while the accuracy drops only by 2.13\% and 2.57\% on average, for CIFAR10 and CIFAR100, respectively. The second key observation that with a larger device-side model, the accuracy of DC-CCL will improves until reaching a critical point. Specifically, with $\alpha_{co}=\frac{1}{4}$, the device-side model size increases by $1.41\times$ and $1.43\times$, and the accuracy improves by 0.77\% and 1.17\% on average, for CIFAR10 and CIFAR100, respectively. Therefore, DC-CCL allows to adjust the size of the device-side model according to available resources, thereby balancing accuracy and efficiency.

%% file: 05RelatedWork.tex
\section{Related Work}

In this section, we review related work and clarify the key difference from DC-CCL.

\subsection{On-Device Inference}

To reduce service latency and communication cost, many models have been offloaded from the cloud to mobile devices for real-time inference (e.g., facial recognition, photo beautification, question answering, and video analytics), no longer requiring to upload the sensitive samples to be predicted. Related work mainly focused on how to enable on-device inference under the resource constraints of mobile devices.

At the algorithm level, different model compression techniques were proposed to reduce model size and optimize model structure, including quantization~\cite{quan,deep_compress}, model pruning~\cite{prune}, knowledge distillation \cite{arxiv15:hinton:knowdist}, and neural architecture search (NAS)~\cite{onceforall, fewshotnas, nasvit}. Mistify~\cite{mistify} further automates the cloud-to-device model porting process, satisfying the customized requirements of heterogeneous mobile devices. At the system level, existing work focused on how to schedule device-side resources. CoDL \cite{codl} splits each layer of a model vertically into subtasks and adaptively allocate them on a CPU and a GPU based on runtime resource conditions. Band \cite{band} partitions the model into a set of operator groups and assigns them to different processors based on their dependencies. Gemel~\cite{gemel} proposed a GPU memory management technique, which exploits architectural similarities among multiple on-device vision models for layer sharing.

The above work involves the cloud-to-device model offloading process and the on-device inference phase, but does not incorporate the on-device training phase and the device-to-cloud knowledge fusion process as in DC-CCL.

\subsection{Efficient Finetuning}

Adapting the model pre-trained on the cloud to diverse downstream tasks is important in practical deployment. Model finetuning over task-specific data is a widely used method. Existing work mainly focused on improving finetuning efficiency, specific to different downstream platforms.

When the downstream tasks are on powerful servers, storing the large full model is allowed, and how to reduce the size of tunable parameters becomes a key problem, especially for large lanange models (e.g., the series of GPT). \citet{lora} proposed LoRA, where the original weights are frozen, and the update is represented with a low-rank decomposition. \citet{bitfit} introduced BitFit, a sparse update method that modifies only the bias terms of the model. \citet{offsite} proposed offsite-tuning, which finetunes the top and bottom layers of model and compresses the large middle layers into a emulator using layer-drop. \citet{proc:emnlp21:prompt} proposed prompt tuning, requiring to store and tune only a small task-specific prompt (i.e., a few tokens) for each downstream task.

When the downstream tasks are on resource-constraint mobile devices, the key problem turns to reducing the size of the entire model. Besides general compression techniques, one line of work developed light-weight convolution kernels specifically for CNNs. \citet{mobilenet} proposed depthwise separable convolution in MobileNets, where the regular convolution is replaced by the combination of a single layer filter and a 1$\times$1 filter. \citet{shufflenet} proposed pointwise group convolution in ShuffleNet, which splits the input channels into groups and applied a 1$\times$1 filter convolution to each group separately. Another line of work further considered reducing training memory. \citet{tinytl} proposed to update only bias term for device-affordable CNNs. \citet{melon} applied memory saving techniques, such as re-computation and micro-batch. \citet{mgemm} designed a novel matrix decomposition and recombination method.

These work, however, did not consider how to transfer the new knowledge from the downstream mobile device to the upstream cloud server, which is the focus of DC-CCL.

\subsection{Device-Cloud Collaborative Learning}

To integrate the natural advantages and the resources of both mobile devices and the cloud in supporting intelligent services, device-cloud collaborative learning (including inference and training) emerged as a new paradigm.  

In device-cloud collaborative inference, existing work considered how to split the inference task between the cloud and the mobile devices, or let the mobile devices perform pre-inference. Neurosurgeon \cite{neurosurgeon} automatically partitions the model computation at the granularity of DNN layers. SPINN \cite{spinn} co-optimises the horizontal splitting of CNN and the early-exit policy to meet user-defined service-level requirements. AgileNN~\cite{agilenn} vertically splits the outputs of the low-level feature extractor, lets the cloud and the mobile device separately complete the remaining forward processes, the outputs of which are combined as the inference results. Reducto \cite{reducto} filters out unnecessary frames on the mobile device using a light-weight model and uploads the other frames to the cloud for video analytics.

In device-cloud collaborative training, according to the interaction mechanism, existing work can be generally divided into three categories: model-based, sample-based, and feature-based. (1) The model-based line of work enables device-cloud collaboration through exchanging model and its update periodically, and requires the model to be deployable on the mobile device. The most typical framework is FL \cite{fed}, which requires the cloud and the mobile devices to use the full model. The follow-up work \cite{niu_mobicom20} focused on the large embedding layer of recommendation or NLP models and proposed to let each mobile device retrieve a few embedding vectors in a key-value way for local training. \cite{hermes} instead focused on CNN, required each mobile device to download the full model for initialization, and introduced structured sparsity regularization into local training, thereby gradually pruning low-magnitude weights. (2) The sample-based line of work allows the mobile devices and the cloud to exchange samples, which is feasible in some application scenarios where data privacy and communication overhead (e.g., the size of each user's behavior data in recommender systems is not very large.) are not major concerns. \citet{yao_kdd21} trained the model over all the users' data and further improved the model performance over long-tail users with knowledge distillation. \citet{yan_kdd22} and \citet{gu_alibaba21} proposed to let each mobile device retrieve some similar samples from the cloud-side global pool, thereby augmenting the local data for personalized training. (3) The feature-based line of work facilitates device-cloud collaborative learning through exchanging the model's intermediate outputs. The typical framework is split learning \cite{splitl}, which offloads the low-level encoder to the mobile device and uploads the low-level features to the cloud. However, the exchange of low-level features may raise privacy concerns and incur high communication overhead. The follow-up work \cite{fedgkt} proposed to alternately optimize the device-side low-level encoder and the cloud-side layers to avoid frequent device-cloud communication.

DC-CCL differs from the above work in several practical considerations, including no large model down-link, no raw sample up-link, and no low-level feature up-link.

\subsection{Decoupled Structure}

In CV scenarios, the deep models adopt dense connections. Some existing work on the cloud-based distributed learning studied model parallelism methods. \citet{modelassemble} proposed to divide the model horizontally into submodels, train them in parallel, and link them for final finetuning. \citet{alexnet} proposed group convolution to facilitate model parallelism, which splits the channels of a convolutional layer into multiple groups and performed convolution within each group, thus decoupling their forward processes. \citet{fed2} later applied group convolution to cross-device FL, decoupling the model into several submodels and linking each submodel with a certain class to mitigate the local training bias caused by label missing. Therefore, if the local samples of a mobile device involve many classes, the number of required submodels is large, breaking the resource limit.

In contrast to CV, the models in recommender systems and NLP normally contain a large and sparse embedding layer for the full set of items/words. A mobile device tends to involve a few items/words, retrieves only the corresponding embedding vectors, and directly takes the other dense network layers \cite{niu_mobicom20}. However, such a decoupling method may be not applicable to other model structures.

Compared with the above work, DC-CCL focuses on CV models, and not only decouples the forward passes of two submodels through vertical splitting but also decouples their training processes via a control model. The model size for the mobile device depends only on the available resource.

%% file: 06Conclusion.tex
\begin{table*}[!t]
    \caption{The CNN architecture for the feasibility study in Section \ref{ftim}.} \label{toy_model}
    \vspace{-0.4em}
    \centering
    \resizebox{0.94\textwidth}{!}{
    \begin{tabular}{cccccccc}
    \toprule
Model Component    &  Layer Name & \#Channels/Neurons & Kernel Size & Stride & Padding & \#Params & Size (MB)\\
    \midrule
    {Shared Encoder}     & conv1 & 3 & 5 & 1 & 2 & 9,600 &  0.04 \\
    \cmidrule(lr){1-8}
    \multirow{6}{*}{Cloud-Side Submodel}    & conv1 & 128 & 3 & 1 & 1 & 258,048 & \multirow{6}{*}{5.37}\\
     & maxpool & 128 & 2 & 2 & / & / \\
     & conv2 & 224 & 3 & 1 & 1 & 451,584\\
     & conv3 & 224 & 5 & 1 & 2 & 627,200 \\
     & maxpool & 224 & 2 & 2 & / & / \\
     & fc1 & 7168 & / & / & / & 71,680\\
     \cmidrule(lr){1-8}
    \multirow{6}{*}{Co-Submodel} & conv1 & 128 & 3 & 1 & 2 & 36,864 & \multirow{6}{*}{0.31}\\
     & maxpool & 128 & 2 & 2 & / & / \\
     & conv2 & 32 & 3 & 1 & 1 & 9,216\\
     & conv3 & 32 & 5 & 1 & 2 & 25,600 \\
     & maxpool & 224 & 2 & 2 & / & / \\
     & fc1 & 1024 & / & / & / & 10,240\\
    \bottomrule
    \end{tabular}
    }
\end{table*}

\begin{table*}[!t]
    \caption{The hyper-parameters used in DC-CCL. \#Cloud Epochs and \#Device Epochs denote the numbers of the local epochs in each round of collaborative learning on the sides of the cloud and the mobile device, respectively.}
    \label{localepochs}
    \centering
    \resizebox{0.95\textwidth}{!}{
    \begin{tabular}{ccccccccccccc}
    \toprule
\multirow{2}{*}{Dataset}     & \multicolumn{2}{c}{Cloud-Side Submodel} & & \multicolumn{2}{c}{Control Model} & & \multicolumn{3}{c}{Co-Submodel} & & \multicolumn{2}{c}{Co-Submodel's Classifier} \\
\cmidrule{2-3}\cmidrule{5-6}\cmidrule{8-10}\cmidrule{12-13}
 & LR & \#Epochs & & LR & \#Epochs & & LR & \#Cloud Epochs & \#Device Epochs & & LR & \#Epochs \\
    \midrule
    {CIFAR10}      & 0.01 & 15 & & 0.001 & 15 & & 0.01 & 1 & 4 & & 0.01 & 10 \\
    \midrule
    {CIFAR100}     & 0.02 & 50 & & 0.002 & 20 & & 0.02 & 1 & 4 & & 0.02 & 10 \\
    \midrule
    {StanfordCars} & 0.01 & 100 & & 0.001 & 100 & & 0.01 & 1 & 4 & & 0.01 & 25 \\
    \midrule
    {UCF101}       & 0.001 & 50 & & 0.0001 & 50 & & 0.001 & 1 & 4 & & 0.01 & 25 \\
    \midrule
    ImageNet-1K \&    & \multirow{2}{*}{0.01} & \multirow{2}{*}{1} & & \multirow{2}{*}{0.001} & \multirow{2}{*}{5} & & \multirow{2}{*}{0.01} & \multirow{2}{*}{1} & \multirow{2}{*}{5} & & \multirow{2}{*}{0.01} & \multirow{2}{*}{1} \\
    StanfordCars & & & & & & & & \\
    \bottomrule
    \end{tabular}
    }
\end{table*}

\section{Conclusion}
In this work, we have proposed DC-CCL, which enables the cloud-side large vision models to continue to learn from fresh device-side samples, given the non-existence of raw sample up-link or large model down-link. DC-CCL vertically splits the base model into a cloud-side submodel and a co-submodel, and only the co-submodel is offloaded to the mobile device for local training with the help of a control model. The evaluation results have demonstrated the effectiveness, efficiency, and superiority of DC-CCL.

%% file: 07Appendix.tex
\appendix

\section{Experimental Details}\label{exp_detail}
The details of the CNN used for feasibility study in Section \ref{ftim} are recorded in Table \ref{toy_model}. The control model takes the same architecture as the co-submodel.

For the experiments in Section \ref{sec_eval}, we search for the optimal the setting of the learning rates and the number of cloud-side local epochs and device-side local epochs per round. The hyper-parameters used in DC-CCL are listed in Table \ref{localepochs}. For fairness in comparison, all the baselines will correspondingly take the same setting as DC-CCL. 

The experimental platform is a workstation with the operating system Ubuntu 18.04.3, CUDA version 11.4, and one NVIDIA GeForce RTX 2080Ti GPU.

\section{Additional Consideration}
In the main body, we propose DC-CCL for the collaborative learning between the cloud and a single mobile device. In fact, the underlying data parallelism framework can be naturally extended to support multiple mobile devices. We need to consider the effect of cross-device data heterogeneity and system heterogeneity on the isomorphism of the device-cloud co-submodel. In particular, the vision samples on different mobile devices are non-iid (e.g., involving different classes), and different mobile devices have heterogeneous resources. We first recall that the size of the device-cloud co-submodel is adjusted according to the resource constraint of the mobile device, not depending on the local samples of the mobile device. As shown in Table \ref{acc_cmp_ideal}, we adopt the same device-side model for the settings of different device-side sample classes. Therefore, the cross-device heterogeneity will not affect the co-submodel. Regarding system heterogeneity, we focus on the container environment of a certain mobile APP, while the runtime capacity of the mobile APP, typically 10MB, has already considered and smoothed the heterogeneity on different types (i.e., low-end, mid-end, and high-end) of mobile devices. If the runtime capacity of the mobile APP varies, DC-CCL also allows scaling up or scaling down the co-submodel with a good performance guarantee, as evaluated in Section \ref{sec:eval:device}.

%% file: main.bbl

\begin{thebibliography}{47}


\ifx \showCODEN    \undefined \def \showCODEN     #1{\unskip}     \fi
\ifx \showDOI      \undefined \def \showDOI       #1{#1}\fi
\ifx \showISBNx    \undefined \def \showISBNx     #1{\unskip}     \fi
\ifx \showISBNxiii \undefined \def \showISBNxiii  #1{\unskip}     \fi
\ifx \showISSN     \undefined \def \showISSN      #1{\unskip}     \fi
\ifx \showLCCN     \undefined \def \showLCCN      #1{\unskip}     \fi
\ifx \shownote     \undefined \def \shownote      #1{#1}          \fi
\ifx \showarticletitle \undefined \def \showarticletitle #1{#1}   \fi
\ifx \showURL      \undefined \def \showURL       {\relax}        \fi
\providecommand\bibfield[2]{#2}
\providecommand\bibinfo[2]{#2}
\providecommand\natexlab[1]{#1}
\providecommand\showeprint[2][]{arXiv:#2}

\bibitem[\protect\citeauthoryear{Cai, Gan, Wang, Zhang, and Han}{Cai
  et~al\mbox{.}}{2020a}]%
        {onceforall}
\bibfield{author}{\bibinfo{person}{Han Cai}, \bibinfo{person}{Chuang Gan},
  \bibinfo{person}{Tianzhe Wang}, \bibinfo{person}{Zhekai Zhang}, {and}
  \bibinfo{person}{Song Han}.} \bibinfo{year}{2020}\natexlab{a}.
\newblock \showarticletitle{Once-for-All: Train One Network and Specialize it
  for Efficient Deployment}. In \bibinfo{booktitle}{\emph{{ICLR}}}.
  \bibinfo{publisher}{OpenReview.net}, \bibinfo{address}{Addis Ababa,
  Ethiopia}.
\newblock


\bibitem[\protect\citeauthoryear{Cai, Gan, Zhu, and Han}{Cai
  et~al\mbox{.}}{2020b}]%
        {tinytl}
\bibfield{author}{\bibinfo{person}{Han Cai}, \bibinfo{person}{Chuang Gan},
  \bibinfo{person}{Ligeng Zhu}, {and} \bibinfo{person}{Song Han}.}
  \bibinfo{year}{2020}\natexlab{b}.
\newblock \showarticletitle{TinyTL: Reduce Memory, Not Parameters for Efficient
  On-Device Learning}. In \bibinfo{booktitle}{\emph{NeurIPS}}.
  \bibinfo{publisher}{MIT press}, \bibinfo{address}{Virtual},
  \bibinfo{pages}{11285--11297}.
\newblock


\bibitem[\protect\citeauthoryear{Deng, Dong, Socher, Li, Li, and Fei-Fei}{Deng
  et~al\mbox{.}}{2009}]%
        {imagenet}
\bibfield{author}{\bibinfo{person}{Jia Deng}, \bibinfo{person}{Wei Dong},
  \bibinfo{person}{Richard Socher}, \bibinfo{person}{Li-Jia Li},
  \bibinfo{person}{Kai Li}, {and} \bibinfo{person}{Li Fei-Fei}.}
  \bibinfo{year}{2009}\natexlab{}.
\newblock \showarticletitle{Imagenet: A large-scale hierarchical image
  database}. In \bibinfo{booktitle}{\emph{CVPR}}. \bibinfo{publisher}{IEEE},
  \bibinfo{address}{Miami, Florida, {USA}}, \bibinfo{pages}{248--255}.
\newblock


\bibitem[\protect\citeauthoryear{Gong, Wang, Li, Chen, Yan, Tian, Liu, and
  Chandra}{Gong et~al\mbox{.}}{2022}]%
        {nasvit}
\bibfield{author}{\bibinfo{person}{Chengyue Gong}, \bibinfo{person}{Dilin
  Wang}, \bibinfo{person}{Meng Li}, \bibinfo{person}{Xinlei Chen},
  \bibinfo{person}{Zhicheng Yan}, \bibinfo{person}{Yuandong Tian},
  \bibinfo{person}{Qiang Liu}, {and} \bibinfo{person}{Vikas Chandra}.}
  \bibinfo{year}{2022}\natexlab{}.
\newblock \showarticletitle{NASViT: Neural Architecture Search for Efficient
  Vision Transformers with Gradient Conflict aware Supernet Training}. In
  \bibinfo{booktitle}{\emph{{ICLR}}}. \bibinfo{publisher}{OpenReview.net},
  \bibinfo{address}{Virtual}.
\newblock


\bibitem[\protect\citeauthoryear{Gu, Niu, Yan, Wu, Tang, Jia, Lyu, and Chen}{Gu
  et~al\mbox{.}}{2022}]%
        {gu_alibaba21}
\bibfield{author}{\bibinfo{person}{Renjie Gu}, \bibinfo{person}{Chaoyue Niu},
  \bibinfo{person}{Yikai Yan}, \bibinfo{person}{Fan Wu},
  \bibinfo{person}{Shaojie Tang}, \bibinfo{person}{Rongfeng Jia},
  \bibinfo{person}{Chengfei Lyu}, {and} \bibinfo{person}{Guihai Chen}.}
  \bibinfo{year}{2022}\natexlab{}.
\newblock \showarticletitle{On-Device Learning with Cloud-Coordinated Data
  Augmentation for Extreme Model Personalization in Recommender Systems}.
\newblock \bibinfo{journal}{\emph{arxiv: 2201.10382}} (\bibinfo{year}{2022}).
\newblock
\urldef\tempurl%
\url{http://arxiv.org/abs/2201.10382}
\showURL{%
\tempurl}


\bibitem[\protect\citeauthoryear{Guo, Hu, and Hu}{Guo et~al\mbox{.}}{2021}]%
        {mistify}
\bibfield{author}{\bibinfo{person}{Peizhen Guo}, \bibinfo{person}{Bo Hu}, {and}
  \bibinfo{person}{Wenjun Hu}.} \bibinfo{year}{2021}\natexlab{}.
\newblock \showarticletitle{Mistify: Automating {DNN} Model Porting for
  On-Device Inference at the Edge}. In \bibinfo{booktitle}{\emph{{NSDI}}}.
  \bibinfo{publisher}{{USENIX}}, \bibinfo{address}{Virtual},
  \bibinfo{pages}{705--719}.
\newblock


\bibitem[\protect\citeauthoryear{Gupta and Raskar}{Gupta and Raskar}{2018}]%
        {splitl}
\bibfield{author}{\bibinfo{person}{Otkrist Gupta} {and} \bibinfo{person}{Ramesh
  Raskar}.} \bibinfo{year}{2018}\natexlab{}.
\newblock \showarticletitle{Distributed learning of deep neural network over
  multiple agents}.
\newblock \bibinfo{journal}{\emph{Journal of Network and Computer
  Applications}}  \bibinfo{volume}{116} (\bibinfo{year}{2018}),
  \bibinfo{pages}{1--8}.
\newblock


\bibitem[\protect\citeauthoryear{Gupta, Agrawal, Gopalakrishnan, and
  Narayanan}{Gupta et~al\mbox{.}}{2015}]%
        {quan}
\bibfield{author}{\bibinfo{person}{Suyog Gupta}, \bibinfo{person}{Ankur
  Agrawal}, \bibinfo{person}{Kailash Gopalakrishnan}, {and}
  \bibinfo{person}{Pritish Narayanan}.} \bibinfo{year}{2015}\natexlab{}.
\newblock \showarticletitle{Deep Learning with Limited Numerical Precision}. In
  \bibinfo{booktitle}{\emph{{ICML}}}. \bibinfo{publisher}{JMLR.org},
  \bibinfo{address}{Lille, France}, \bibinfo{pages}{1737--1746}.
\newblock


\bibitem[\protect\citeauthoryear{Han, Mao, and Dally}{Han
  et~al\mbox{.}}{2016}]%
        {deep_compress}
\bibfield{author}{\bibinfo{person}{Song Han}, \bibinfo{person}{Huizi Mao},
  {and} \bibinfo{person}{William~J. Dally}.} \bibinfo{year}{2016}\natexlab{}.
\newblock \showarticletitle{Deep Compression: Compressing Deep Neural Network
  with Pruning, Trained Quantization and Huffman Coding}. In
  \bibinfo{booktitle}{\emph{{ICLR}}}. \bibinfo{publisher}{OpenReview.net},
  \bibinfo{address}{San Juan, Puerto Rico}.
\newblock


\bibitem[\protect\citeauthoryear{Han, Pool, Tran, and Dally}{Han
  et~al\mbox{.}}{2015}]%
        {prune}
\bibfield{author}{\bibinfo{person}{Song Han}, \bibinfo{person}{Jeff Pool},
  \bibinfo{person}{John Tran}, {and} \bibinfo{person}{William~J. Dally}.}
  \bibinfo{year}{2015}\natexlab{}.
\newblock \showarticletitle{Learning both Weights and Connections for Efficient
  Neural Network}. In \bibinfo{booktitle}{\emph{{NeurIPS}}}.
  \bibinfo{publisher}{MIT press}, \bibinfo{address}{Montreal, Quebec, Canada},
  \bibinfo{pages}{1135--1143}.
\newblock


\bibitem[\protect\citeauthoryear{He, Annavaram, and Avestimehr}{He
  et~al\mbox{.}}{2020}]%
        {fedgkt}
\bibfield{author}{\bibinfo{person}{Chaoyang He}, \bibinfo{person}{Murali
  Annavaram}, {and} \bibinfo{person}{Salman Avestimehr}.}
  \bibinfo{year}{2020}\natexlab{}.
\newblock \showarticletitle{Group Knowledge Transfer: Federated Learning of
  Large CNNs at the Edge}. In \bibinfo{booktitle}{\emph{NeurIPS}}.
  \bibinfo{publisher}{MIT press}, \bibinfo{address}{Virtual},
  \bibinfo{pages}{14068--14080}.
\newblock


\bibitem[\protect\citeauthoryear{He, Zhang, Ren, and Sun}{He
  et~al\mbox{.}}{2016}]%
        {resnet}
\bibfield{author}{\bibinfo{person}{Kaiming He}, \bibinfo{person}{Xiangyu
  Zhang}, \bibinfo{person}{Shaoqing Ren}, {and} \bibinfo{person}{Jian Sun}.}
  \bibinfo{year}{2016}\natexlab{}.
\newblock \showarticletitle{Deep residual learning for image recognition}. In
  \bibinfo{booktitle}{\emph{CVPR}}. \bibinfo{publisher}{IEEE},
  \bibinfo{address}{Las Vegas, NV, USA}, \bibinfo{pages}{770--778}.
\newblock


\bibitem[\protect\citeauthoryear{Hinton, Vinyals, and Dean}{Hinton
  et~al\mbox{.}}{2015}]%
        {arxiv15:hinton:knowdist}
\bibfield{author}{\bibinfo{person}{Geoffrey~E. Hinton}, \bibinfo{person}{Oriol
  Vinyals}, {and} \bibinfo{person}{Jeffrey Dean}.}
  \bibinfo{year}{2015}\natexlab{}.
\newblock \showarticletitle{Distilling the Knowledge in a Neural Network}.
\newblock \bibinfo{journal}{\emph{arXiv: 1503.02531}} (\bibinfo{year}{2015}).
\newblock
\urldef\tempurl%
\url{http://arxiv.org/abs/1503.02531}
\showURL{%
\tempurl}


\bibitem[\protect\citeauthoryear{Hong, Yao, Zhou, Zhang, and Wang}{Hong
  et~al\mbox{.}}{2023}]%
        {lt2}
\bibfield{author}{\bibinfo{person}{Feng Hong}, \bibinfo{person}{Jiangchao Yao},
  \bibinfo{person}{Zhihan Zhou}, \bibinfo{person}{Ya Zhang}, {and}
  \bibinfo{person}{Yanfeng Wang}.} \bibinfo{year}{2023}\natexlab{}.
\newblock \showarticletitle{Long-Tailed Partial Label Learning via Dynamic
  Rebalancing}. In \bibinfo{booktitle}{\emph{{ICLR}}}.
  \bibinfo{publisher}{OpenReview.net}, \bibinfo{address}{Kigali, Rwanda}.
\newblock


\bibitem[\protect\citeauthoryear{Howard, Sandler, Chu, Chen, Chen, Tan, Wang,
  Zhu, Pang, Vasudevan, Le, and Adam}{Howard et~al\mbox{.}}{2019}]%
        {mobilenetv3-small}
\bibfield{author}{\bibinfo{person}{Andrew Howard}, \bibinfo{person}{Mark
  Sandler}, \bibinfo{person}{Grace Chu}, \bibinfo{person}{Liang-Chieh Chen},
  \bibinfo{person}{Bo Chen}, \bibinfo{person}{Mingxing Tan},
  \bibinfo{person}{Weijun Wang}, \bibinfo{person}{Yukun Zhu},
  \bibinfo{person}{Ruoming Pang}, \bibinfo{person}{Vijay Vasudevan},
  \bibinfo{person}{Quoc~V. Le}, {and} \bibinfo{person}{Hartwig Adam}.}
  \bibinfo{year}{2019}\natexlab{}.
\newblock \showarticletitle{Searching for MobileNetV3}. In
  \bibinfo{booktitle}{\emph{ICCV}}. \bibinfo{publisher}{IEEE},
  \bibinfo{address}{Seoul, Korea (South)}, \bibinfo{pages}{1314--1324}.
\newblock


\bibitem[\protect\citeauthoryear{Howard, Zhu, Chen, Kalenichenko, Wang, Weyand,
  Andreetto, and Adam}{Howard et~al\mbox{.}}{2017}]%
        {mobilenet}
\bibfield{author}{\bibinfo{person}{Andrew~G. Howard}, \bibinfo{person}{Menglong
  Zhu}, \bibinfo{person}{Bo Chen}, \bibinfo{person}{Dmitry Kalenichenko},
  \bibinfo{person}{Weijun Wang}, \bibinfo{person}{Tobias Weyand},
  \bibinfo{person}{Marco Andreetto}, {and} \bibinfo{person}{Hartwig Adam}.}
  \bibinfo{year}{2017}\natexlab{}.
\newblock \showarticletitle{MobileNets: Efficient Convolutional Neural Networks
  for Mobile Vision Applications}.
\newblock \bibinfo{journal}{\emph{arXiv: 1704.04861}} (\bibinfo{year}{2017}).
\newblock
\urldef\tempurl%
\url{http://arxiv.org/abs/1704.04861}
\showURL{%
\tempurl}


\bibitem[\protect\citeauthoryear{Hu, Shen, Wallis, Allen{-}Zhu, Li, Wang, Wang,
  and Chen}{Hu et~al\mbox{.}}{2022a}]%
        {lora}
\bibfield{author}{\bibinfo{person}{Edward~J. Hu}, \bibinfo{person}{Yelong
  Shen}, \bibinfo{person}{Phillip Wallis}, \bibinfo{person}{Zeyuan
  Allen{-}Zhu}, \bibinfo{person}{Yuanzhi Li}, \bibinfo{person}{Shean Wang},
  \bibinfo{person}{Lu Wang}, {and} \bibinfo{person}{Weizhu Chen}.}
  \bibinfo{year}{2022}\natexlab{a}.
\newblock \showarticletitle{LoRA: Low-Rank Adaptation of Large Language
  Models}. In \bibinfo{booktitle}{\emph{{ICLR}}}.
  \bibinfo{publisher}{OpenReview.net}, \bibinfo{address}{Virtual}.
\newblock


\bibitem[\protect\citeauthoryear{Hu, Wang, Hong, Li, Hsieh, and Feng}{Hu
  et~al\mbox{.}}{2022b}]%
        {fewshotnas}
\bibfield{author}{\bibinfo{person}{Shoukang Hu}, \bibinfo{person}{Ruochen
  Wang}, \bibinfo{person}{Lanqing Hong}, \bibinfo{person}{Zhenguo Li},
  \bibinfo{person}{Cho{-}Jui Hsieh}, {and} \bibinfo{person}{Jiashi Feng}.}
  \bibinfo{year}{2022}\natexlab{b}.
\newblock \showarticletitle{Generalizing Few-Shot {NAS} with Gradient
  Matching}. In \bibinfo{booktitle}{\emph{{ICLR}}}.
  \bibinfo{publisher}{OpenReview.net}, \bibinfo{address}{Virtual}.
\newblock


\bibitem[\protect\citeauthoryear{Huang and Gao}{Huang and Gao}{2022}]%
        {agilenn}
\bibfield{author}{\bibinfo{person}{Kai Huang} {and} \bibinfo{person}{Wei Gao}.}
  \bibinfo{year}{2022}\natexlab{}.
\newblock \showarticletitle{Real-time neural network inference on extremely
  weak devices: agile offloading with explainable {AI}}. In
  \bibinfo{booktitle}{\emph{MobiCom}}. \bibinfo{publisher}{{ACM}},
  \bibinfo{address}{Sydney, NSW, Australia}, \bibinfo{pages}{200--213}.
\newblock


\bibitem[\protect\citeauthoryear{Jeong, Lee, Kim, Jeon, Jeong, Lee, and
  Chun}{Jeong et~al\mbox{.}}{2022}]%
        {band}
\bibfield{author}{\bibinfo{person}{Joo~Seong Jeong}, \bibinfo{person}{Jingyu
  Lee}, \bibinfo{person}{Donghyun Kim}, \bibinfo{person}{Changmin Jeon},
  \bibinfo{person}{Changjin Jeong}, \bibinfo{person}{Youngki Lee}, {and}
  \bibinfo{person}{Byung{-}Gon Chun}.} \bibinfo{year}{2022}\natexlab{}.
\newblock \showarticletitle{Band: coordinated multi-DNN inference on
  heterogeneous mobile processors}. In \bibinfo{booktitle}{\emph{MobiSys}}.
  \bibinfo{publisher}{{ACM}}, \bibinfo{address}{Portland, Oregon},
  \bibinfo{pages}{235--247}.
\newblock


\bibitem[\protect\citeauthoryear{Jia, Zhang, Cao, Jiang, Liu, Ren, and
  Zhang}{Jia et~al\mbox{.}}{2022}]%
        {codl}
\bibfield{author}{\bibinfo{person}{Fucheng Jia}, \bibinfo{person}{Deyu Zhang},
  \bibinfo{person}{Ting Cao}, \bibinfo{person}{Shiqi Jiang},
  \bibinfo{person}{Yunxin Liu}, \bibinfo{person}{Ju Ren}, {and}
  \bibinfo{person}{Yaoxue Zhang}.} \bibinfo{year}{2022}\natexlab{}.
\newblock \showarticletitle{CoDL: efficient {CPU-GPU} co-execution for deep
  learning inference on mobile devices}. In
  \bibinfo{booktitle}{\emph{MobiSys}}. \bibinfo{publisher}{{ACM}},
  \bibinfo{address}{Portland, Oregon}, \bibinfo{pages}{209--221}.
\newblock


\bibitem[\protect\citeauthoryear{Kang, Xie, Rohrbach, Yan, Gordo, Feng, and
  Kalantidis}{Kang et~al\mbox{.}}{2020}]%
        {lt1}
\bibfield{author}{\bibinfo{person}{Bingyi Kang}, \bibinfo{person}{Saining Xie},
  \bibinfo{person}{Marcus Rohrbach}, \bibinfo{person}{Zhicheng Yan},
  \bibinfo{person}{Albert Gordo}, \bibinfo{person}{Jiashi Feng}, {and}
  \bibinfo{person}{Yannis Kalantidis}.} \bibinfo{year}{2020}\natexlab{}.
\newblock \showarticletitle{Decoupling Representation and Classifier for
  Long-Tailed Recognition}. In \bibinfo{booktitle}{\emph{{ICLR}}}.
  \bibinfo{publisher}{OpenReview.net}, \bibinfo{address}{Addis Ababa,
  Ethiopia}.
\newblock


\bibitem[\protect\citeauthoryear{Kang, Hauswald, Gao, Rovinski, Mudge, Mars,
  and Tang}{Kang et~al\mbox{.}}{2017}]%
        {neurosurgeon}
\bibfield{author}{\bibinfo{person}{Yiping Kang}, \bibinfo{person}{Johann
  Hauswald}, \bibinfo{person}{Cao Gao}, \bibinfo{person}{Austin Rovinski},
  \bibinfo{person}{Trevor~N. Mudge}, \bibinfo{person}{Jason Mars}, {and}
  \bibinfo{person}{Lingjia Tang}.} \bibinfo{year}{2017}\natexlab{}.
\newblock \showarticletitle{Neurosurgeon: Collaborative Intelligence Between
  the Cloud and Mobile Edge}. In \bibinfo{booktitle}{\emph{{ASPLOS}}}.
  \bibinfo{publisher}{{ACM}}, \bibinfo{address}{Xi'an, China},
  \bibinfo{pages}{615--629}.
\newblock


\bibitem[\protect\citeauthoryear{Krause, Jin, Yang, and Fei-Fei}{Krause
  et~al\mbox{.}}{2015}]%
        {cars_data}
\bibfield{author}{\bibinfo{person}{Jonathan Krause}, \bibinfo{person}{Hailin
  Jin}, \bibinfo{person}{Jianchao Yang}, {and} \bibinfo{person}{Li Fei-Fei}.}
  \bibinfo{year}{2015}\natexlab{}.
\newblock \showarticletitle{Fine-grained recognition without part annotations}.
  In \bibinfo{booktitle}{\emph{CVPR}}. \bibinfo{publisher}{IEEE},
  \bibinfo{address}{Boston, MA, USA}, \bibinfo{pages}{5546--5555}.
\newblock


\bibitem[\protect\citeauthoryear{Krizhevsky, Hinton, et~al\mbox{.}}{Krizhevsky
  et~al\mbox{.}}{2009}]%
        {cifar}
\bibfield{author}{\bibinfo{person}{Alex Krizhevsky}, \bibinfo{person}{Geoffrey
  Hinton}, {et~al\mbox{.}}} \bibinfo{year}{2009}\natexlab{}.
\newblock \showarticletitle{Learning multiple layers of features from tiny
  images}.
\newblock  (\bibinfo{year}{2009}).
\newblock


\bibitem[\protect\citeauthoryear{Krizhevsky, Sutskever, and Hinton}{Krizhevsky
  et~al\mbox{.}}{2012}]%
        {alexnet}
\bibfield{author}{\bibinfo{person}{Alex Krizhevsky}, \bibinfo{person}{Ilya
  Sutskever}, {and} \bibinfo{person}{Geoffrey~E. Hinton}.}
  \bibinfo{year}{2012}\natexlab{}.
\newblock \showarticletitle{ImageNet Classification with Deep Convolutional
  Neural Networks}. In \bibinfo{booktitle}{\emph{{NeurIPS}}}.
  \bibinfo{publisher}{MIT press}, \bibinfo{address}{Lake Tahoe, Nevada, United
  States}, \bibinfo{pages}{1106--1114}.
\newblock


\bibitem[\protect\citeauthoryear{Laskaridis, Venieris, Almeida, Leontiadis, and
  Lane}{Laskaridis et~al\mbox{.}}{2020}]%
        {spinn}
\bibfield{author}{\bibinfo{person}{Stefanos Laskaridis},
  \bibinfo{person}{Stylianos~I. Venieris}, \bibinfo{person}{M{\'{a}}rio
  Almeida}, \bibinfo{person}{Ilias Leontiadis}, {and}
  \bibinfo{person}{Nicholas~D. Lane}.} \bibinfo{year}{2020}\natexlab{}.
\newblock \showarticletitle{{SPINN:} synergistic progressive inference of
  neural networks over device and cloud}. In
  \bibinfo{booktitle}{\emph{MobiCom}}. \bibinfo{publisher}{{ACM}},
  \bibinfo{address}{London, United Kingdom}, \bibinfo{pages}{37:1--37:15}.
\newblock


\bibitem[\protect\citeauthoryear{Lester, Al{-}Rfou, and Constant}{Lester
  et~al\mbox{.}}{2021}]%
        {proc:emnlp21:prompt}
\bibfield{author}{\bibinfo{person}{Brian Lester}, \bibinfo{person}{Rami
  Al{-}Rfou}, {and} \bibinfo{person}{Noah Constant}.}
  \bibinfo{year}{2021}\natexlab{}.
\newblock \showarticletitle{The Power of Scale for Parameter-Efficient Prompt
  Tuning}. In \bibinfo{booktitle}{\emph{{EMNLP}}}. \bibinfo{publisher}{ACL},
  \bibinfo{address}{Punta Cana, Dominican Republic},
  \bibinfo{pages}{3045--3059}.
\newblock


\bibitem[\protect\citeauthoryear{Li, Sun, Li, Pu, Li, and Chen}{Li
  et~al\mbox{.}}{2021}]%
        {hermes}
\bibfield{author}{\bibinfo{person}{Ang Li}, \bibinfo{person}{Jingwei Sun},
  \bibinfo{person}{Pengcheng Li}, \bibinfo{person}{Yu Pu}, \bibinfo{person}{Hai
  Li}, {and} \bibinfo{person}{Yiran Chen}.} \bibinfo{year}{2021}\natexlab{}.
\newblock \showarticletitle{Hermes: an efficient federated learning framework
  for heterogeneous mobile clients}. In \bibinfo{booktitle}{\emph{MobiCom}}.
  \bibinfo{publisher}{{ACM}}, \bibinfo{address}{New Orleans, Louisiana, USA},
  \bibinfo{pages}{420--437}.
\newblock


\bibitem[\protect\citeauthoryear{Li, Padmanabhan, Zhao, Wang, Xu, and
  Netravali}{Li et~al\mbox{.}}{2020}]%
        {reducto}
\bibfield{author}{\bibinfo{person}{Yuanqi Li}, \bibinfo{person}{Arthi
  Padmanabhan}, \bibinfo{person}{Pengzhan Zhao}, \bibinfo{person}{Yufei Wang},
  \bibinfo{person}{Guoqing~Harry Xu}, {and} \bibinfo{person}{Ravi Netravali}.}
  \bibinfo{year}{2020}\natexlab{}.
\newblock \showarticletitle{Reducto: On-Camera Filtering for Resource-Efficient
  Real-Time Video Analytics}. In \bibinfo{booktitle}{\emph{{SIGCOMM}}}.
  \bibinfo{publisher}{{ACM}}, \bibinfo{address}{Virtual},
  \bibinfo{pages}{359--376}.
\newblock


\bibitem[\protect\citeauthoryear{Lv, Niu, Gu, Jiang, Wang, Liu, Wu, Yao, Huang,
  Huang, Huang, Shu, Song, Zou, Lan, Xu, Wu, Tang, Wu, and Chen}{Lv
  et~al\mbox{.}}{2022}]%
        {proc:osdi22:walle}
\bibfield{author}{\bibinfo{person}{Chengfei Lv}, \bibinfo{person}{Chaoyue Niu},
  \bibinfo{person}{Renjie Gu}, \bibinfo{person}{Xiaotang Jiang},
  \bibinfo{person}{Zhaode Wang}, \bibinfo{person}{Bin Liu},
  \bibinfo{person}{Ziqi Wu}, \bibinfo{person}{Qiulin Yao},
  \bibinfo{person}{Congyu Huang}, \bibinfo{person}{Panos Huang},
  \bibinfo{person}{Tao Huang}, \bibinfo{person}{Hui Shu},
  \bibinfo{person}{Jinde Song}, \bibinfo{person}{Bin Zou},
  \bibinfo{person}{Peng Lan}, \bibinfo{person}{Guohuan Xu},
  \bibinfo{person}{Fei Wu}, \bibinfo{person}{Shaojie Tang},
  \bibinfo{person}{Fan Wu}, {and} \bibinfo{person}{Guihai Chen}.}
  \bibinfo{year}{2022}\natexlab{}.
\newblock \showarticletitle{Walle: An {End-to-End}, {General-Purpose}, and
  {Large-Scale} Production System for {Device-Cloud} Collaborative Machine
  Learning}. In \bibinfo{booktitle}{\emph{{OSDI}}}.
  \bibinfo{publisher}{USENIX}, \bibinfo{address}{Carlsbad, CA, USA},
  \bibinfo{pages}{249--265}.
\newblock


\bibitem[\protect\citeauthoryear{McMahan, Moore, Ramage, Hampson, and
  y~Arcas}{McMahan et~al\mbox{.}}{2017}]%
        {fed}
\bibfield{author}{\bibinfo{person}{Brendan McMahan}, \bibinfo{person}{Eider
  Moore}, \bibinfo{person}{Daniel Ramage}, \bibinfo{person}{Seth Hampson},
  {and} \bibinfo{person}{Blaise~Ag{\"{u}}era y Arcas}.}
  \bibinfo{year}{2017}\natexlab{}.
\newblock \showarticletitle{Communication-Efficient Learning of Deep Networks
  from Decentralized Data}. In \bibinfo{booktitle}{\emph{AISTATS}}.
  \bibinfo{publisher}{{JMLR}}, \bibinfo{address}{Fort Lauderdale, {USA}},
  \bibinfo{pages}{1273--1282}.
\newblock


\bibitem[\protect\citeauthoryear{Ni, Wang, Yu, Jiang, Cao, and Huang}{Ni
  et~al\mbox{.}}{2022}]%
        {modelassemble}
\bibfield{author}{\bibinfo{person}{Zanlin Ni}, \bibinfo{person}{Yulin Wang},
  \bibinfo{person}{Jiangwei Yu}, \bibinfo{person}{Haojun Jiang},
  \bibinfo{person}{Yue Cao}, {and} \bibinfo{person}{Gao Huang}.}
  \bibinfo{year}{2022}\natexlab{}.
\newblock \showarticletitle{Deep Model Assembling}.
\newblock \bibinfo{journal}{\emph{arXiv: 2212.04129}} (\bibinfo{year}{2022}).
\newblock
\urldef\tempurl%
\url{http://arxiv.org/abs/2212.04129}
\showURL{%
\tempurl}


\bibitem[\protect\citeauthoryear{Niu, Wu, Tang, Hua, Jia, Lv, Wu, and Chen}{Niu
  et~al\mbox{.}}{2020}]%
        {niu_mobicom20}
\bibfield{author}{\bibinfo{person}{Chaoyue Niu}, \bibinfo{person}{Fan Wu},
  \bibinfo{person}{Shaojie Tang}, \bibinfo{person}{Lifeng Hua},
  \bibinfo{person}{Rongfei Jia}, \bibinfo{person}{Chengfei Lv},
  \bibinfo{person}{Zhihua Wu}, {and} \bibinfo{person}{Guihai Chen}.}
  \bibinfo{year}{2020}\natexlab{}.
\newblock \showarticletitle{Billion-scale federated learning on mobile clients:
  a submodel design with tunable privacy}. In
  \bibinfo{booktitle}{\emph{MobiCom}}. \bibinfo{publisher}{{ACM}},
  \bibinfo{address}{London, United Kingdom}, \bibinfo{pages}{31:1--31:14}.
\newblock


\bibitem[\protect\citeauthoryear{Padmanabhan, Agarwal, Iyer, Ananthanarayanan,
  Shu, Karianakis, Xu, and Netravali}{Padmanabhan et~al\mbox{.}}{2023}]%
        {gemel}
\bibfield{author}{\bibinfo{person}{Arthi Padmanabhan}, \bibinfo{person}{Neil
  Agarwal}, \bibinfo{person}{Anand~P. Iyer}, \bibinfo{person}{Ganesh
  Ananthanarayanan}, \bibinfo{person}{Yuanchao Shu}, \bibinfo{person}{Nikolaos
  Karianakis}, \bibinfo{person}{Guoqing~Harry Xu}, {and} \bibinfo{person}{Ravi
  Netravali}.} \bibinfo{year}{2023}\natexlab{}.
\newblock \showarticletitle{{GEMEL:} Model Merging for Memory-Efficient,
  Real-Time Video Analytics at the Edge}. In \bibinfo{booktitle}{\emph{NSDI}}.
  \bibinfo{publisher}{{USENIX}}, \bibinfo{address}{Boston, MA, USA}.
\newblock


\bibitem[\protect\citeauthoryear{Park, Bin, and Lee}{Park
  et~al\mbox{.}}{2022}]%
        {mgemm}
\bibfield{author}{\bibinfo{person}{Jongseok Park}, \bibinfo{person}{Kyungmin
  Bin}, {and} \bibinfo{person}{Kyunghan Lee}.} \bibinfo{year}{2022}\natexlab{}.
\newblock \showarticletitle{mGEMM: low-latency convolution with minimal memory
  overhead optimized for mobile devices}. In
  \bibinfo{booktitle}{\emph{MobiSys}}. \bibinfo{publisher}{{ACM}},
  \bibinfo{address}{Portland, Oregon}, \bibinfo{pages}{222--234}.
\newblock


\bibitem[\protect\citeauthoryear{Simonyan and Zisserman}{Simonyan and
  Zisserman}{2014}]%
        {vgg}
\bibfield{author}{\bibinfo{person}{Karen Simonyan} {and}
  \bibinfo{person}{Andrew Zisserman}.} \bibinfo{year}{2014}\natexlab{}.
\newblock \showarticletitle{Very deep convolutional networks for large-scale
  image recognition}.
\newblock \bibinfo{journal}{\emph{arXiv:1409.1556}} (\bibinfo{year}{2014}).
\newblock
\urldef\tempurl%
\url{http://arxiv.org/abs/1409.1556}
\showURL{%
\tempurl}


\bibitem[\protect\citeauthoryear{Soomro, Zamir, and Shah}{Soomro
  et~al\mbox{.}}{2012}]%
        {ucf101}
\bibfield{author}{\bibinfo{person}{Khurram Soomro},
  \bibinfo{person}{Amir~Roshan Zamir}, {and} \bibinfo{person}{Mubarak Shah}.}
  \bibinfo{year}{2012}\natexlab{}.
\newblock \showarticletitle{UCF101: A dataset of 101 human actions classes from
  videos in the wild}.
\newblock \bibinfo{journal}{\emph{arXiv:1212.0402}} (\bibinfo{year}{2012}).
\newblock
\urldef\tempurl%
\url{http://arxiv.org/abs/1212.0402}
\showURL{%
\tempurl}


\bibitem[\protect\citeauthoryear{Tan and Le}{Tan and Le}{2019}]%
        {efficientnet}
\bibfield{author}{\bibinfo{person}{Mingxing Tan} {and} \bibinfo{person}{Quoc
  Le}.} \bibinfo{year}{2019}\natexlab{}.
\newblock \showarticletitle{Efficientnet: Rethinking model scaling for
  convolutional neural networks}. In \bibinfo{booktitle}{\emph{ICML}}.
  \bibinfo{publisher}{JMLR.org}, \bibinfo{address}{Long Beach, California,
  {USA}}, \bibinfo{pages}{6105--6114}.
\newblock


\bibitem[\protect\citeauthoryear{Tran, Bourdev, Fergus, Torresani, and
  Paluri}{Tran et~al\mbox{.}}{2015}]%
        {c3d}
\bibfield{author}{\bibinfo{person}{Du Tran}, \bibinfo{person}{Lubomir Bourdev},
  \bibinfo{person}{Rob Fergus}, \bibinfo{person}{Lorenzo Torresani}, {and}
  \bibinfo{person}{Manohar Paluri}.} \bibinfo{year}{2015}\natexlab{}.
\newblock \showarticletitle{Learning Spatiotemporal Features With 3D
  Convolutional Networks}. In \bibinfo{booktitle}{\emph{ICCV}}.
  \bibinfo{publisher}{IEEE}, \bibinfo{address}{Santiago, Chile},
  \bibinfo{pages}{4489--4497}.
\newblock


\bibitem[\protect\citeauthoryear{Wang, Xu, Jin, Dong, Yuan, Jin, Huang, Liu,
  and Liu}{Wang et~al\mbox{.}}{2022}]%
        {melon}
\bibfield{author}{\bibinfo{person}{Qipeng Wang}, \bibinfo{person}{Mengwei Xu},
  \bibinfo{person}{Chao Jin}, \bibinfo{person}{Xinran Dong},
  \bibinfo{person}{Jinliang Yuan}, \bibinfo{person}{Xin Jin},
  \bibinfo{person}{Gang Huang}, \bibinfo{person}{Yunxin Liu}, {and}
  \bibinfo{person}{Xuanzhe Liu}.} \bibinfo{year}{2022}\natexlab{}.
\newblock \showarticletitle{Melon: breaking the memory wall for
  resource-efficient on-device machine learning}. In
  \bibinfo{booktitle}{\emph{MobiSys}}. \bibinfo{publisher}{{ACM}},
  \bibinfo{address}{Portland, Oregon}, \bibinfo{pages}{450--463}.
\newblock


\bibitem[\protect\citeauthoryear{Xiao, Lin, and Han}{Xiao
  et~al\mbox{.}}{2023}]%
        {offsite}
\bibfield{author}{\bibinfo{person}{Guangxuan Xiao}, \bibinfo{person}{Ji Lin},
  {and} \bibinfo{person}{Song Han}.} \bibinfo{year}{2023}\natexlab{}.
\newblock \showarticletitle{Offsite-Tuning: Transfer Learning without Full
  Model}.
\newblock \bibinfo{journal}{\emph{arXiv: 2302.04870}} (\bibinfo{year}{2023}).
\newblock
\urldef\tempurl%
\url{http://arxiv.org/abs/2302.04870}
\showURL{%
\tempurl}


\bibitem[\protect\citeauthoryear{Yan, Niu, Gu, Wu, Tang, Hua, Lyu, and
  Chen}{Yan et~al\mbox{.}}{2022}]%
        {yan_kdd22}
\bibfield{author}{\bibinfo{person}{Yikai Yan}, \bibinfo{person}{Chaoyue Niu},
  \bibinfo{person}{Renjie Gu}, \bibinfo{person}{Fan Wu},
  \bibinfo{person}{Shaojie Tang}, \bibinfo{person}{Lifeng Hua},
  \bibinfo{person}{Chengfei Lyu}, {and} \bibinfo{person}{Guihai Chen}.}
  \bibinfo{year}{2022}\natexlab{}.
\newblock \showarticletitle{On-Device Learning for Model Personalization with
  Large-Scale Cloud-Coordinated Domain Adaption}. In
  \bibinfo{booktitle}{\emph{{KDD}}}. \bibinfo{publisher}{{ACM}},
  \bibinfo{address}{Washington, DC, USA}, \bibinfo{pages}{2180--2190}.
\newblock


\bibitem[\protect\citeauthoryear{Yao, Wang, Jia, Han, Zhou, and Yang}{Yao
  et~al\mbox{.}}{2021}]%
        {yao_kdd21}
\bibfield{author}{\bibinfo{person}{Jiangchao Yao}, \bibinfo{person}{Feng Wang},
  \bibinfo{person}{Kunyang Jia}, \bibinfo{person}{Bo Han},
  \bibinfo{person}{Jingren Zhou}, {and} \bibinfo{person}{Hongxia Yang}.}
  \bibinfo{year}{2021}\natexlab{}.
\newblock \showarticletitle{Device-Cloud Collaborative Learning for
  Recommendation}. In \bibinfo{booktitle}{\emph{{KDD}}}.
  \bibinfo{publisher}{{ACM}}, \bibinfo{address}{Virtual},
  \bibinfo{pages}{3865--3874}.
\newblock


\bibitem[\protect\citeauthoryear{Yu, Zhang, Qin, Xu, Wang, Liu, Tian, and
  Chen}{Yu et~al\mbox{.}}{2021}]%
        {fed2}
\bibfield{author}{\bibinfo{person}{Fuxun Yu}, \bibinfo{person}{Weishan Zhang},
  \bibinfo{person}{Zhuwei Qin}, \bibinfo{person}{Zirui Xu}, \bibinfo{person}{Di
  Wang}, \bibinfo{person}{Chenchen Liu}, \bibinfo{person}{Zhi Tian}, {and}
  \bibinfo{person}{Xiang Chen}.} \bibinfo{year}{2021}\natexlab{}.
\newblock \showarticletitle{Fed2: Feature-Aligned Federated Learning}. In
  \bibinfo{booktitle}{\emph{{KDD}}}. \bibinfo{publisher}{{ACM}},
  \bibinfo{address}{Virtual}, \bibinfo{pages}{3865--3874}.
\newblock


\bibitem[\protect\citeauthoryear{Zaken, Goldberg, and Ravfogel}{Zaken
  et~al\mbox{.}}{2022}]%
        {bitfit}
\bibfield{author}{\bibinfo{person}{Elad~Ben Zaken}, \bibinfo{person}{Yoav
  Goldberg}, {and} \bibinfo{person}{Shauli Ravfogel}.}
  \bibinfo{year}{2022}\natexlab{}.
\newblock \showarticletitle{BitFit: Simple Parameter-efficient Fine-tuning for
  Transformer-based Masked Language-models}. In
  \bibinfo{booktitle}{\emph{{ACL}}}. \bibinfo{publisher}{Association for
  Computational Linguistics}, \bibinfo{address}{Dublin, Ireland},
  \bibinfo{pages}{1--9}.
\newblock


\bibitem[\protect\citeauthoryear{Zhang, Zhou, Lin, and Sun}{Zhang
  et~al\mbox{.}}{2018}]%
        {shufflenet}
\bibfield{author}{\bibinfo{person}{Xiangyu Zhang}, \bibinfo{person}{Xinyu
  Zhou}, \bibinfo{person}{Mengxiao Lin}, {and} \bibinfo{person}{Jian Sun}.}
  \bibinfo{year}{2018}\natexlab{}.
\newblock \showarticletitle{ShuffleNet: An Extremely Efficient Convolutional
  Neural Network for Mobile Devices}. In \bibinfo{booktitle}{\emph{{CVPR}}}.
  \bibinfo{publisher}{Computer Vision Foundation / {IEEE} Computer Society},
  \bibinfo{address}{Salt Lake City, UT, USA}, \bibinfo{pages}{6848--6856}.
\newblock


\end{thebibliography}
